\documentclass{article}

\usepackage{microtype}
\usepackage{graphicx}
\usepackage{subfigure}
\usepackage{booktabs} 

\usepackage{hyperref}

\usepackage[accepted]{icml2024}

\usepackage{amsmath}
\usepackage{amssymb}
\usepackage{mathtools}
\usepackage{amsthm}

\usepackage{url}
\usepackage{multirow}
\usepackage{tcolorbox}
\usepackage{enumitem}
\usepackage[T1]{fontenc}
\usepackage{inconsolata}
\usepackage{subcaption}

\usepackage[capitalize,noabbrev]{cleveref}

\theoremstyle{plain}

\theoremstyle{definition}

\theoremstyle{remark}

\usepackage[textsize=tiny]{todonotes}

\icmltitlerunning{Language Agents with Reinforcement Learning for Strategic Play in the Werewolf Game}

\begin{document}

\twocolumn[
\icmltitle{Language Agents with Reinforcement Learning for \\ Strategic Play in the Werewolf Game}



\icmlsetsymbol{equal}{*}

\begin{icmlauthorlist}
\icmlauthor{Zelai Xu}{thu}
\icmlauthor{Chao Yu}{thu}
\icmlauthor{Fei Fang}{cmu}
\icmlauthor{Yu Wang}{equal,thu}
\icmlauthor{Yi Wu}{equal,thu,sqz}
\end{icmlauthorlist}

\icmlaffiliation{thu}{Tsinghua University, Beijing, China}
\icmlaffiliation{cmu}{Carnegie Mellon University, Pittsburgh, United States}
\icmlaffiliation{sqz}{Shanghai Qi Zhi Institute, Shanghai, China}

\icmlcorrespondingauthor{Zelai Xu}{zelai.eecs@gmail.com}
\icmlcorrespondingauthor{Yi Wu}{jxwuyi@gmail.com}

\icmlkeywords{AI Agents, Large Language Models, Reinforcement Learning, Social Deduction Games}

\vskip 0.3in
]



\printAffiliationsAndNotice{\icmlEqualContribution} 

\begin{abstract}
Agents built with large language models (LLMs) have shown great potential across a wide range of domains. However, in complex decision-making tasks, pure LLM-based agents tend to exhibit intrinsic bias in their choice of actions, which is inherited from the model's training data and results in suboptimal performance. To develop \textit{strategic language agents}, i.e., agents that generate flexible language actions and possess strong decision-making abilities, we propose a novel framework that powers LLM-based agents with reinforcement learning (RL). We consider Werewolf, a popular social deduction game, as a challenging testbed that emphasizes versatile communication and strategic gameplay. To mitigate the intrinsic bias in language actions, our agents use an LLM to perform deductive reasoning and generate a diverse set of action candidates. Then an RL policy trained to optimize the decision-making ability chooses an action from the candidates to play in the game. Extensive experiments show that our agents overcome the intrinsic bias and outperform existing LLM-based agents in the Werewolf game. We also conduct human-agent experiments and find that our agents achieve human-level performance and demonstrate strong strategic play.
\end{abstract}

\section{Introduction}\label{sec:intro}
\begin{figure}[t]
    \centering
    \includegraphics[width=\linewidth]{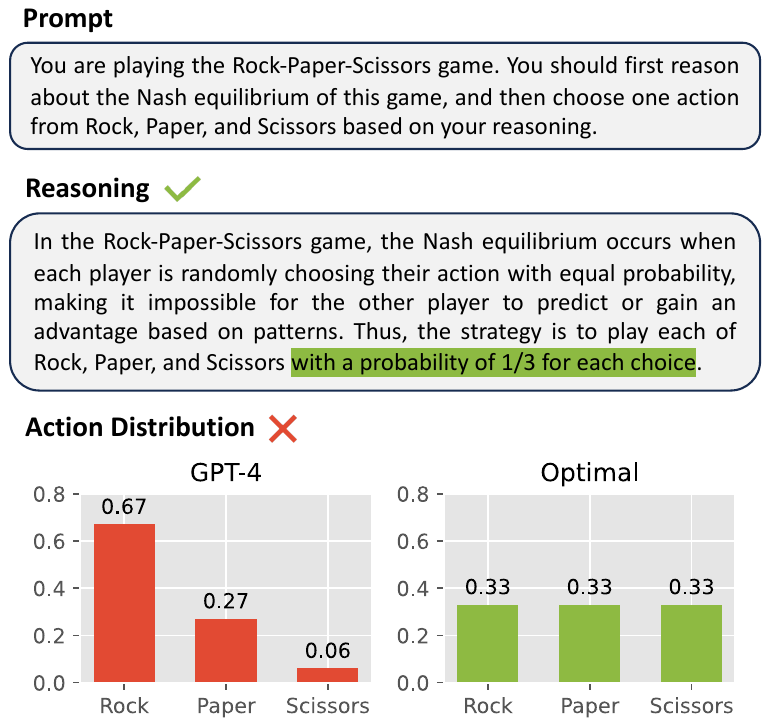}
    \caption{The intrinsic bias in the action distribution of pure LLM-based agents despite the correct reasoning result.}
    \label{fig:bias}
\end{figure}

Building agents that are capable of logical reasoning, strategic decision-making, and communication with humans has been a longstanding aspiration in the field of AI~\citep{wooldridge1995intelligent,goodwin1995formalizing}.
With their extensive knowledge and emergent generalizability, large language models (LLMs) have shown great potential in constructing intelligent agents and have led to many recent advancements~\citep{ouyang2022training,wei2022emergent}.
These LLM-based agents demonstrate proficiency in a wide range of tasks including web surfing~\citep{nakano2021webgpt,yao2022react,zheng2024gpt}, complex video games~\citep{wang2023voyager,ma2023large}, and real-world applications~\citep{ahn2022can,shah2023lm}.
Moreover, in multi-agent scenarios, LLM-based agents exhibit the ability to produce human-like interactions~\citep{park2023generative,williams2023epidemic}, achieve zero-shot cooperation~\citep{li2023camel,chen2023agentverse}, and compete with adversarial players~\citep{meta2022human,wang2023avalon}.

Despite the impressive achievements, we observe that in complex decision-making tasks such as multi-agent games, pure LLM-based agents built with prompting techniques tend to exhibit intrinsic bias in their choice of actions, resulting in suboptimal performance.
Consider a simple example where we use GPT-4 to play the Rock-Paper-Scissors game.
To generate the optimal strategy, we prompt the LLM to first reason about the Nash equilibrium of this game and then choose an action based on the reasoning result.
We let this pure LLM-based agent play 100 independent games and profile its reasoning result and action distribution in Fig~\ref{fig:bias}.
In all 100 games, the agent successfully identifies the optimal strategy as randomly choosing an action.
However, even with this perfect reasoning result, the agent's actual action distribution shows a clear bias toward choosing Rock, which can be easily exploited by an opponent who always plays Paper.
This intrinsic bias is inevitably inherited from the model's pretraining data and hinders pure LLM-based agents from strong strategic play in more complex scenarios.
A recent work, Cicero~\citep{meta2022human}, addresses this issue in the game of Diplomacy by learning a policy to choose from predefined actions that specify the agents' intents and then generating action-conditioned dialogues with a language model.
Nevertheless, the action space of their policy is a fixed, finite set of game-specific actions, while real-world interactions between humans often feature free-form conversations with nuanced intents, which induces an unbounded action space for agents to play.

In this work, we aim to develop \textit{strategic language agents}, i.e., agents that generate flexible language actions and possess strong decision-making abilities.
We consider the Werewolf game as a challenging mixed cooperative-competitive multi-agent testbed.
Werewolf is one of the most popular social deduction games where two teams of players with hidden roles interact in natural language to uncover and defeat their opponents.
During gameplay, the Werewolves need to lie about their roles and secretly eliminate other players, while the Villagers need to share information and vote out the hidden Werewolves.
The game's high demand for language proficiency necessitates the use of LLMs to build intelligent agents.
However, pure LLM-based agents face the challenges of deducing hidden roles with deceptive information and overcoming intrinsic bias in language actions, which result in unsatisfactory performance.

To build strategic language agents for the Werewolf game, we propose a novel framework that powers LLM-based agents with reinforcement learning (RL) to address these challenges.
Our framework consists of three components.
The first component is hidden role deduction, which uses an LLM to distinguish between truths and lies and explicitly deduce the hidden role of each player to facilitate subsequent decision-making.
The second component is diverse action generation, which prompts the LLM to generate a diverse set of action candidates instead of a single action to mitigate the intrinsic bias.
The last component is population-based RL training, which learns an RL policy to optimize the distribution over the action candidates and improves policy robustness by playing against a population of various agents.
The combination of the LLM and the RL policy enables our agents to deduce hidden information and overcome intrinsic bias, leading to strong strategic play in the Werewolf game.

We conduct extensive experiments to evaluate the performance of strategic language agents in Werewolf and demonstrate the benefit of our design.
We first perform case studies and visualize the action distributions to validate that our agents overcome the intrinsic bias and learn better policy.
Then we compare our agents with four state-of-the-art baselines in a round-robin tournament where our agents achieve the highest win rate against all agents.
Moreover, we let our agents play with human players and find they achieve comparable win rates to average humans as both teammates and opponents.
We also conduct systematic ablations to show the effectiveness of the key components in our framework.

\section{Related Work}\label{sec:related}
\begin{figure*}[t]
    \centering
    \includegraphics[width=\textwidth]{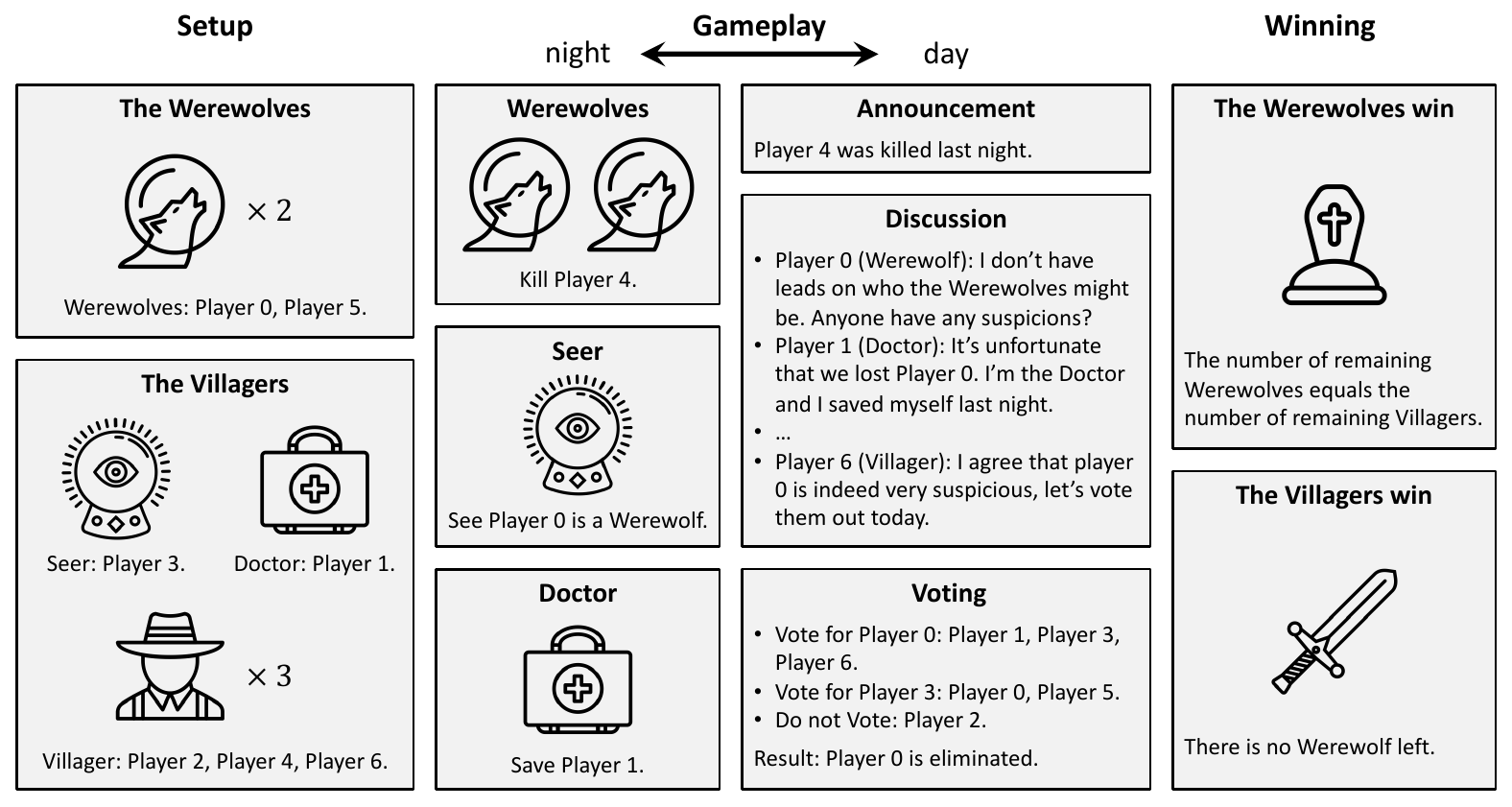}
    \caption{An example of the Werewolf game. Players are randomly assigned a hidden role and divided into the Werewolves and the Villagers. The game alternates between night and day rounds until one team achieves the winning condition.}
    \label{fig:game}
\end{figure*}

\textbf{Building Agents with Large Language Models.}
There is a recent trend in developing agents with large language models for various domains including website scenarios~\citep{nakano2021webgpt,yao2022webshop,deng2023mind2web}, game and simulation~\citep{huang2022language,wang2023describe,wang2023voyager,ma2023large}, real-world embodiment~\citep{ahn2022can,huang2022inner,vemprala2023chatgpt}, and multi-agent environments~\citep{park2023generative,li2023camel,chen2023agentverse}.
A shared foundation of these works is to utilize the reasoning and in-context learning ability of LLMs to facilitate decision-making.
Chain-of-Thought (CoT)~\citep{wei2022chain} is perhaps the most well-known work that unlocks the reasoning ability of LLMs by asking them to think step-by-step.
To synergize reasoning and acting, ReAct~\citep{yao2022react} uses LLMs to generate both reasoning traces and action plans and proves effective in various benchmarks.
Follow-up work continues to improve performance by incorporating self-reflection~\citep{shinn2023reflexion} and strategic reasoning~\citep{gandhi2023strategic}.
However, even with the perfect reasoning results, these pure LLM-based agents could suffer from the model's intrinsic bias and make suboptimal decisions in more complex scenarios.
Our work takes a step further to address this issue by using the LLM to generate a diverse set of action candidates with minimal bias and training an RL policy to optimize decision-making.

Some other works also augment LLMs with external modules to enhance the agents' decision-making ability.
One representative work is Cicero~\citep{meta2022human} which combines LLMs with RL and achieves human-level play in the game of Diplomacy.
The main difference from our work is that their policy can choose from a fixed finite set of game-specific actions because Diplomacy is a board game with negotiation.
By contrast, Werewolf is a language game without predefined actions, requiring our agents to generate language actions in an unbounded space.
Another work is Tree-of-Thought (ToT)~\citep{yao2023tree} which generalizes CoT by producing multiple thoughts at each reasoning step to create a tree structure and search for optimal plans.
The extensive tree search could be inefficient in complex environments with prohibitively large or even infinite game trees.
LLM+P~\citep{liu2023llm+} incorporate classic planners by converting natural language task descriptions into planning domain definition language (PDDL).
However, PDDL is limited to classic planning problems and cannot be applied to non-cooperative language games like Werewolf.

\textbf{Reinforcement Learning in Non-Cooperative Games.}
Applying reinforcement learning to non-cooperative games has achieved great success in the game of Go~\citep{silver2016mastering,silver2018general}, poker~\citep{moravvcik2017deepstack,brown2018superhuman,brown2019superhuman}, and video games~\citep{vinyals2019grandmaster,berner2019dota}.
The most popular method underlying these achievements is self-play and its variants~\citep{heinrich2015fictitious,heinrich2016deep,hennes2020neural,xu2023fictitious}, which learn a policy by training against itself and its past checkpoints.
Population-based training (PBT) methods like policy-space response oracles (PSRO)~\citep{lanctot2017unified,muller2019generalized} and league training~\citep{vinyals2019grandmaster} generalize self-play by maintaining a pool of various policies and training against the population. 
Another notable line of work is based on regret minimization techniques such as counterfactual regret minimization (CFR)~\citep{zinkevich2007regret,lanctot2009monte,tammelin2014solving,brown2019deep} that iteratively refine the strategies by minimizing the difference between actual outcomes and potential better choices.
Most prior works apply these methods to typical RL environments with vectorized actions, while we combine LLM with RL to generate language actions and learn in the Werewolf game.

\textbf{AI Agents for Social Deduction Games.}
Social deduction games like Werewolf require both strong communication skills and strategic decision-making.
Earlier agents~\citep{wang2018application} lack language proficiency and usually rely on predefined protocols instead of natural language to play a simplified game.
DeepRole~\citep{serrino2019finding} integrates deductive reasoning into CFR to play another social deduction game named Avalon without communication.
With the presence of LLMs, more agents have been developed to play these games in natural language.
A concurrent study~\citep{xu2023exploring} builds a Werewolf agent by heuristic information retrieval and past experience reflection.
Another work proposes ReCon~\citep{wang2023avalon} to play Avalon by thinking from both the agent's and the opponents' perspective.
Despite their deliberate designs, these pure LLM-based agents are not strong enough in decision-making as they still face the issue of intrinsic bias. 
We use them as baselines to compare with our agents in the experiments.

\section{The Werewolf Game}\label{sec:game}
We aim to build strong agents for the seven-player Werewolf game with two Werewolves, one Seer, one Doctor, and three Villagers. 
An example of the game is shown in Fig.~\ref{fig:game} and the detailed rules can be found in Appendix~\ref{app:rules}.

\subsection{Text-Based Environemnt}

We implement a pure text-based Werewolf environment that does not consider external factors like the players' tone or facial expressions.
The observations and actions are text in natural language and the game proceeds as follows.

\textbf{Setup.}
At the beginning of the game, each player is randomly assigned a hidden role, which divides them into the Werewolves and the Villagers (Seer, Doctor, and Villagers).
The two Werewolves know each other's role and hence also know which players are the Villagers.
Their goal is to ``kill'' the other players without being discovered.
The Villagers do not know the hidden roles of other players and their goal is to identify and eliminate the Werewolves.

\textbf{Gameplay.}
The game alternates between night and day rounds, starting with the night. 
In the night round, everyone closes their eyes to let the Werewolves, Seer, and Doctor take secret actions.
The Werewolves choose one player to kill, the Seer chooses one player to check if they are a Werewolf, and the Doctor chooses one player to save without knowing who is the target of the Werewolves.
If the Doctor chooses the same player as the Werewolves, the player is successfully saved and no one is killed in the night round.
Otherwise, the player is eliminated from the game.

In the day round, an announcement is first made to every player about who was killed or no one was killed last night.
Then the remaining players take turns to speak in a discussion about who might be the Werewolves.
Players can claim or lie about their roles, share or withhold information they have discovered, and accuse or defend other players to achieve their purposes.
After all players have participated in the discussion, a vote is held to eliminate one suspicious player.
Each player can vote for one player or choose not to vote, and the player with the most votes is eliminated.
Then the game continues to the next night round until the Werewolves or the Villagers win the game.

\textbf{Winning.}
The Werewolves win if the number of remaining Werewolves is equal to the number of remaining Villagers.
The Villagers win if both Werewolves are eliminated.

The observation of each player is a list of information in natural language that describes the current game history.
This includes their ID, hidden role, secret actions at night (if any) as well as the announcements, discussions, and voting results in the day rounds.
Werewolves also have their teammate's ID and secret actions in the observation.
The players' actions are also in natural language and can be divided into three types.
The first type is secret actions at night including killing, seeing, and saving that choose a specific player as the target.
The second type is statement actions during discussion that share information and convey the player's opinion.
The last type is voting actions that vote for one surviving player or choose not to vote.

\subsection{Challenges for LLM-Based Agents}

Werewolf is a challenging mixed cooperative-competitive game that involves deceptive communication and strategic gameplay.
Its high demand for language proficiency necessitates the use of LLMs to build intelligent agents.
We identify two major challenges of Werewolf that pose higher requirements on the ability of LLM-based agents.

The first challenge is to deduce hidden information from deceptive communication.
With the existence of unknown opponents, agents could be misled by manipulative statements and fail to make accurate judgments about other players.
This requires the agents to distinguish between truths and lies and deduce the hidden roles of other players.

The second challenge is to overcome the intrinsic bias in LLMs and optimize decision-making.
As illustrated by the Rock-Paper-Scissor example in Fig~\ref{fig:bias}, pure LLM-based agents could have suboptimal biases in their action distributions, hindering their ability to make stronger decisions.
This issue becomes more pronounced in competitive games like Werewolf because opponents can exploit the biased distributions to predict the agents' behaviors and counteract to gain advantages.
Therefore, agents must overcome intrinsic bias to reduce exploitation and achieve strong performance.

\section{Strategic Language Agents}\label{sec:method}
\begin{figure*}[t]
    \centering
    \includegraphics[width=\textwidth]{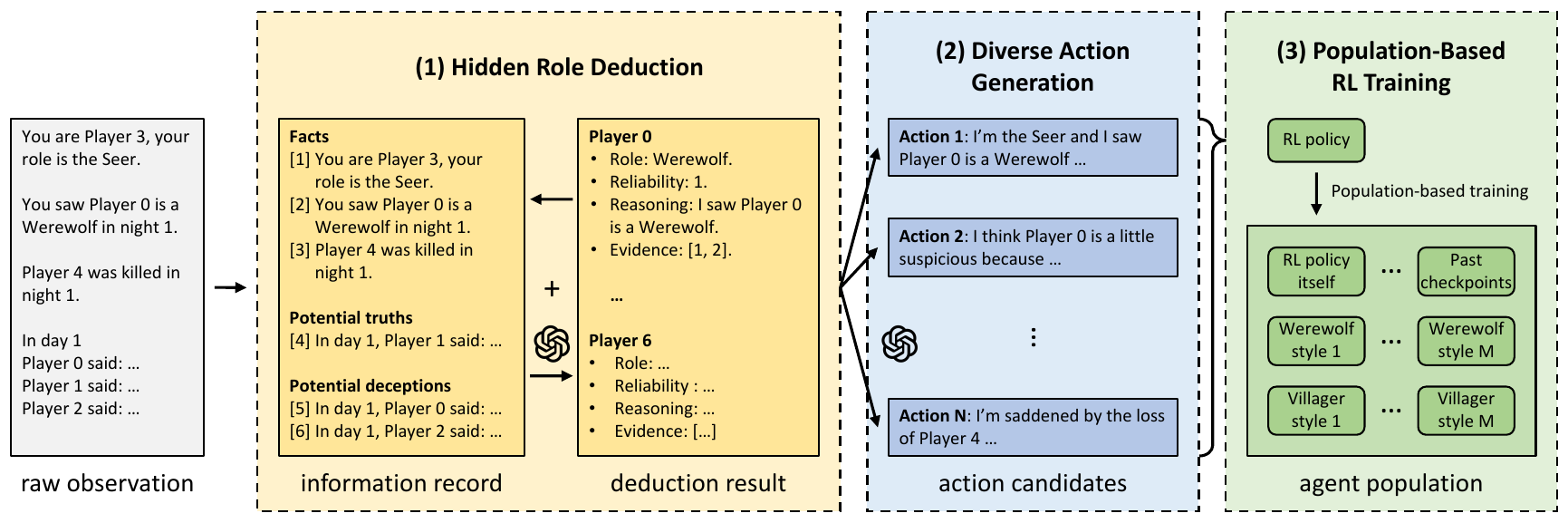}
    \caption{Overview of strategic language agents. (1) Hidden role deduction: categorize key information and deduce hidden roles with the LLM to facilitate subsequent decision-making. (2) Diverse action generation: prompt the LLM for a diverse set of action candidates to mitigate intrinsic bias. (3) Population-based RL training: learn a separate RL policy to optimize the distribution over action candidates and enhance decision-making by playing against a population of various agents.}
    \label{fig:overview}
\end{figure*}

To address the aforementioned challenges, we propose a novel framework that powers LLM-based agents with RL training to build agents that generate flexible language actions and possess strong decision-making ability, which we call \textit{strategic language agents}.
As shown in Fig.~\ref{fig:overview}, our agents consist of three components: hidden role deduction, diverse action generation, and population-based RL training.

\subsection{Hidden Role Deduction}\label{sec:method:reasoning}

The first component aims to improve our agents' deductive reasoning ability in the presence of deceptive information.
we use an LLM to convert the raw observation into an organized information record and a structured deduction result.
The information record extracts key information and distinguishes between truthful and deceptive statements, while the deduction result infers the hidden role of each player and rates their reliability.
These structured and informative data serve as inputs for the next two components and lay the foundation for subsequent decision-making.

Concretely, the information record categorizes the raw observation, which is a list of information in natural language, into three types including facts, potential truths, and potential deceptions.
All information except the statements in discussion is categorized as facts, which encompass the player's role, secret actions, announcements, and voting results.
The statements are categorized according to the speaker's reliability in the deduction result.
In our implementation, the reliability is rated on a scale from 1 to 10.
If a player's reliability is larger than 6, their statements are regarded as potential truths, otherwise as potential deceptions.

With the organized information record, we then prompt the LLM to deduce the hidden roles of other players.
For each player, the LLM is asked to generate four attributes including role, reliability, reasoning, and evidence.
The role attribute infers the most likely hidden role of the player and the reliability attribute rates the credibility of their statements from 1 to 10.
The reasoning attribute is an auxiliary one that explicitly shows the deduction process.
The last evidence attribute is a list of integers that cite items from the information record to support the current deduction.
The evidences are used to extract key information that contributes to the deduction of hidden roles.
If a statement is never cited as evidence, it is regarded as an uninformative item and removed from the information record.
More detailed designs and prompts can be found in Appendix~\ref{app:prompts}.

\subsection{Diverse Action Generation}\label{sec:method:prompting}

To mitigate the intrinsic bias of LLMs and reduce exploitation from adversarial players, agents should be able to consider a broader range of reasonable actions instead of a single biased action.
We propose the diverse action generation component to prompt the LLM for a set of action candidates with strategic diversity.
More specifically, given the information record and deduction result as input, we consider two approaches to generate $N$ action candidates.

\textbf{Vanilla Prompting.}
The first approach generates all action candidates in a single inference by asking the LLM to propose $N$ diverse action candidates that correspond to different strategies.
This approach is simple yet effective for easier actions like secret actions and voting actions, which only need to choose a player as the target.

\textbf{Iterative Prompting.}
The second approach prompts the LLM for $N$ times and iteratively generates one action candidate at a time.
In each inference, we add the existing action candidates to the input and ask the LLM to propose a new action candidate that is strategically different from existing ones.
By having more interactions with the LLM, this approach is empirically found to produce better performance for more complex actions like statement actions.

In our implementation, we use the first approach to generate secret actions and vote actions for fewer inferences, and the second approach to generate statement actions for higher quality.
We also ask the LLM to output their reasoning for each action candidate for interpretability and better performance.
The detailed prompts can be found in Appendix~\ref{app:prompts}.

\begin{figure*}[t]
\centering
\subfigure[Werewolf first night action.]
{
    \centering
    \includegraphics[scale=0.63,trim={0.16cm 0.36cm 0.08cm 0.2cm},clip]{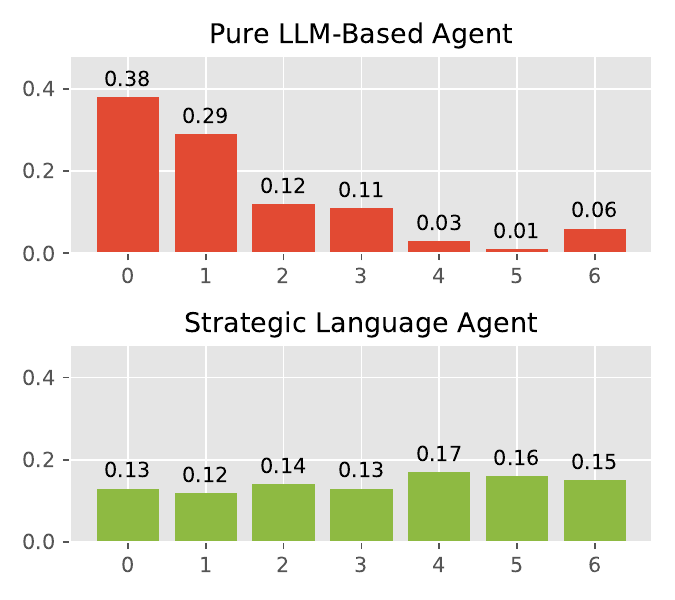}
    \label{fig:werewolf_action}
}
\hfill
\subfigure[Doctor first night action.]
{
    \centering
    \includegraphics[scale=0.63,trim={0.16cm 0.36cm 0.08cm 0.2cm},clip]{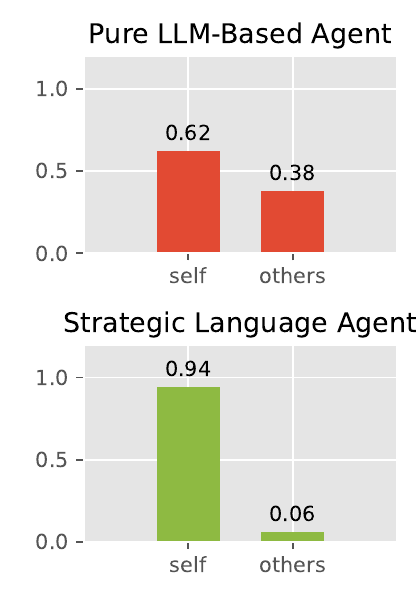}
    \label{fig:doctor_action}
}
\hfill
\subfigure[Villager voting action.]
{
    \centering
    \includegraphics[scale=0.63,trim={0.16cm 0.36cm 0.08cm 0.2cm},clip]{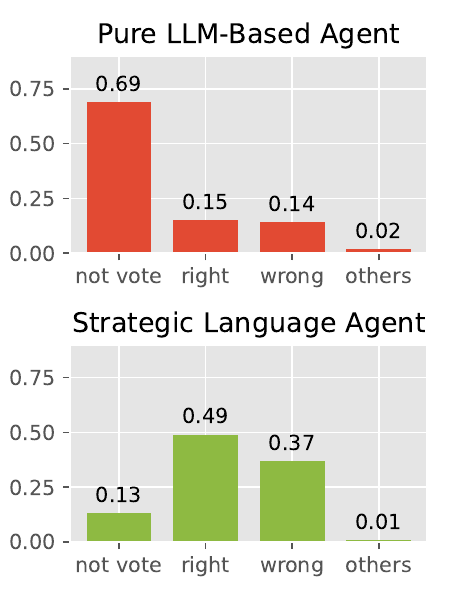}
    \label{fig:villager_action}
}
\caption{Comparison of the action distributions of the pure LLM-based agent and our strategic language agent.}
\label{fig:action_dist}
\end{figure*}

\subsection{Population-Based RL Training}\label{sec:method:training}

With the diverse set of action candidates, agents can avoid fixed patterns and choose from a variety of actions to take.
Although random sampling already leads to unpredictable play, the optimal policy in most cases is a non-uniform distribution over the action candidates.
Directly fine-tuning the LLM to refine the distribution requires a substantial amount of data and computation.
We take an alternative approach that efficiently learns a smaller RL policy to optimize the action distribution and enhance decision-making ability by playing against a population of various agents.

The main difference between our setting and classic RL environments is that the action space of our policy is a set of language actions generated during the game.
Because the action space is not predefined, we cannot use typical policy networks that only take the observation as input and produce a distribution over the fixed action set.
Instead, we adopt a self-attention network~\citep{vaswani2017attention} that takes the observation and all action candidates as input to produce the action distribution.
Concretely, the observation and action candidates in natural language are converted to vector embeddings using the LLM embedding API.
We also use a vector to include player-specific information like their ID, role, etc. and pass it through an encoder to produce a player embedding of the same dimension.
The player embedding, observation embedding, and action candidate embeddings are passed through a residual self-attention block, and the probability to sample an action candidate is proportional to the scaled dot-product attention between the observation embedding and action candidate embedding.

To achieve strong and robust performance with various teammates and opponents, we train the RL policy with a population of diverse agents.
This is because real-world games are usually non-transitive~\citep{czarnecki2020real}, i.e., the policies may cycle like Rock-Paper-Scissors, and population-based training can help agents get over the endless cycles and achieve a higher level of play.
Our agent population consists of two types of policies.
The first type is the RL policy itself and its past checkpoints.
Training with these policies resembles self-play and fictitious play.
The second type is pure LLM-based agents with predefined play styles, which introduce more strategic diversity.
These agents only have the first hidden role deduction component and generate one action according to their predefined styles.
In our implementation, we consider three common styles for the Werewolves and three styles for the Villagers.
More implementation and training details can be found in Appendix~\ref{app:impl}.

\section{Experiments}\label{sec:exp}
We conduct experiments from four different aspects to comprehensively evaluate the performance of strategic language agents.
We first perform case studies and visualize the action distributions of our agents to check if they address the intrinsic bias issue.
Then we compare our agents with four existing LLM-based agents in a round-robin tournament to assess their decision-making ability.
Furthermore, we conduct human-agent experiments to evaluate their performance with and against human players.
We also perform ablation studies to investigate the contribution of the key components in our design.
More experiments, discussions, and emergent behaviors can be found in Appendix~\ref{app:eval}, \ref{app:discussion}, \ref{app:behaviors}.

\subsection{Case Studies}\label{sec:exp:case}

To validate that our method overcomes the intrinsic bias in LLMs,
we investigate three typical situations in the Werewolf game and compare the action distributions of our agents with a pure LLM-based agent that only has the hidden role deduction component and follows \textit{ReAct}~\citep{yao2022react} to generate one single action to play.

\textbf{Werewolf First Night Action.}
On the first night, the Werewolves need to choose a player to kill without any information.
Their optimal policy is to randomly choose a player other than themselves so that the Doctor cannot predict and save the victim.
However, as shown in Fig.~\ref{fig:werewolf_action}, the pure LLM-based agent has a clear bias toward killing Player 0, which could be exploited by a Doctor who always saves Player 0.
By contrast, our agents produce an almost uniform distribution which is close to the optimal policy.

\textbf{Doctor First Night Action.} 
The Doctor also needs to choose a player to save without knowing who is the victim on the first night.
From the Doctor's perspective, any player other than themselves could be a Werewolf, and saving others may waste the action on a Werewolf.
Therefore, the optimal policy for the Doctor is to always save themselves on the first night.
As shown in Fig.~\ref{fig:doctor_action}, the pure LLM-based agent exhibits a suboptimal policy that saves others with a probability of $0.38$, while our agents learn to save themselves with a high probability of $0.94$.

\textbf{Villager Voting Action with Two Self-Proclaimed Seers.}
Another common situation is when two players claim to be the Seer and the Villagers need to choose their actions in the voting phase.
The pure LLM-based agent tends to make conservative decisions and chooses not to vote with a high probability of $0.69$, as shown in Fig~\ref{fig:villager_action}.
However, one of the self-proclaimed Seers must be a Werewolf and not voting for any player only makes it easier for the Werewolves to control the voting result and eliminate the real Seer.
In comparison, our agents have a much lower probability for choosing not to vote and learn to identify and vote out the right Werewolf who pretends to be the Seer.

\begin{figure}[t]
    \centering
    \includegraphics[width=\linewidth,trim={0.4cm 0.4cm 0.4cm 0.4cm},clip]{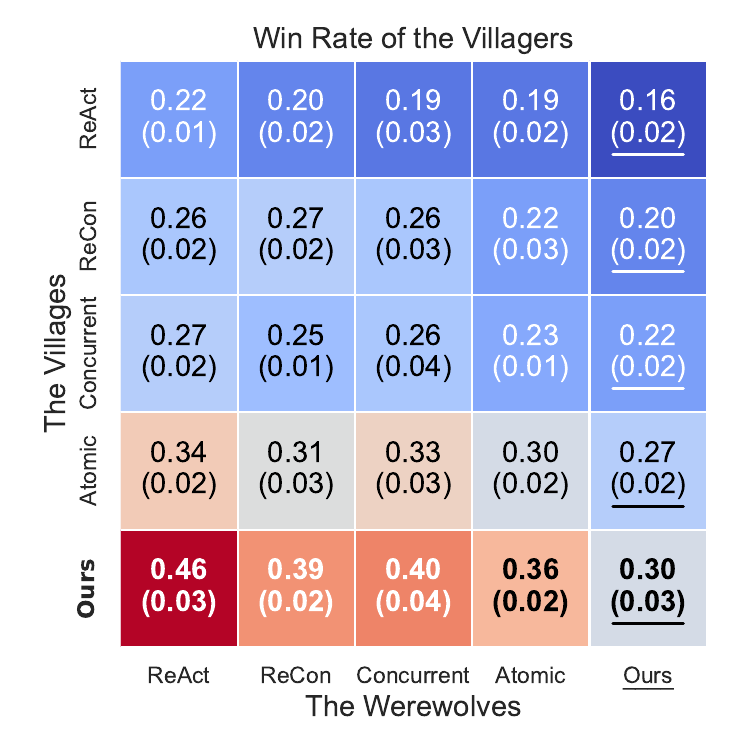}
    \caption{Win rate matrix of the round-robin tournament.}
    \label{fig:tournament}
\end{figure}

\subsection{Round-Robin Tournament}\label{sec:exp:tournament}

We evaluate the performance of our agents in the Werewolf game by playing against existing LLM-based agents.
Because there is only one agent~\citep{xu2023exploring} that works out of the box for Werewolf, we make our best effort to select four representative methods as baselines and make them compatible with our environment. 
We implement the \textit{ReAct}~\citep{yao2022react} agent by directly prompting the LLM with the raw observation to generate both reasoning and action.
\textit{ReCon}~\citep{wang2023avalon} is originally designed for Avalon and generates actions by planning from their own perspective and refining the plan from others' perspective.
The \textit{Concurrent} agent~\citep{xu2023exploring} considers a different version of the Werewolf game and employs heuristic information retrieval and reflection on past experiences.
We make necessary changes to the prompts of \textit{ReCon} and \textit{Concurrent} to apply them in our setting.
The \textit{Atomic} agent is inspired by Cicero~\citep{meta2022human} and has a predefined set of $13$ atomic actions like ``target Player 0'', ``claim to be the Seer'', etc.
This agent learns an RL policy to select an atomic action based on the raw observation and then prompts the LLM to follow the atomic action and generate the language action used in actual gameplay. 

We let our agents and these four baseline agents participate in a round-robin tournament that runs head-to-head competitions between each pair of agents.
For a pair of agents (\textit{A}, \textit{B}), we run $100$ Werewolf games with five \textit{A} agents being the Villagers and two \textit{B} agents being the Werewolves.
This leads to a $5 \times 5$ cross-play matrix in Fig.~\ref{fig:tournament} that records the win rates of the Villagers.
For example, the value in the first row and last column corresponds to the win rate of \textit{ReAct} when it plays as the Villagers against our agents as the Werewolves.
The row in the matrix represents the agent's performance as the Villagers against different opponents, and a row with higher (red) values means stronger performance as the Villagers.
As shown by the bold numbers in the last row, our agents achieve the highest win rates as the Villagers against all agents.
Similarly, the column represents the agent's performance as the Werewolves and smaller (blue) values mean lower win rates of their opponents, indicating stronger performance as the Werewolves.
The last column with underlined numbers shows that our agents also achieve the best performance as the Werewolves.

The strong decision-making ability of our agents comes from the combination of an LLM to generate flexible language actions and an RL policy to overcome intrinsic bias.
The \textit{React}, \textit{ReCon}, and \textit{Concurrent} agents directly generate language actions with an LLM.
Even if the latter two have specific designs for social deduction games, they still face the intrinsic bias issue and have unsatisfactory performance.
On the other hand, although the \textit{Atomic} agent also combines an LLM with an RL policy, its language generation is conditioned on the predefined action set which is hard, if not impossible, to cover the unbounded language space.
More examples and analysis can be found in Appendix~\ref{app:discussion}.

\begin{table*}[t]
\centering
\begin{tabular}{ccccccc}
\toprule
\multirow{2}{*}{Win Rate } & \multicolumn{3}{c}{Play with Six Copies of Our Agents} & \multicolumn{3}{c}{Play with Six Human Players} \\
& Pure LLM-Based & \textbf{Ours} & Human & Pure LLM-Based & \textbf{Ours} & \textbf{Human} \\ 
\midrule
As the Villagers  & 0.25 (0.03) & \textbf{0.30 (0.03)} & 0.28 (0.06) & 0.29 (0.06) & 0.36 (0.05) & \textbf{0.38 (0.07)} \\
As the Werewolves & 0.61 (0.05) & \textbf{0.70 (0.03)} & 0.67 (0.08) & 0.54 (0.08) & \textbf{0.67 (0.06)} & 0.66 (0.08) \\
Overall           & 0.42 (0.04) & \textbf{0.50 (0.03)} & 0.48 (0.07) & 0.42 (0.07) & \textbf{0.52 (0.06)} & \textbf{0.52 (0.08)} \\
\bottomrule
\end{tabular}

\caption{Win rate of one AI agent or human when playing with six copies of our agents and playing with six human players.}
\label{tab:human}
\vspace{-2mm}
\end{table*}

\begin{table}[t]
\centering
\begin{tabular}{ccc}
\toprule
Win Rate & As the Villagers & As the Werewolves \\
\midrule
\textbf{Ours} & \textbf{0.30 (0.03)} & \textbf{0.70 (0.03)} \\
- RL Policy   & 0.23 (0.03) & 0.61 (0.04) \\
- Diversity   & 0.22 (0.04) & 0.61 (0.05) \\
- Deduction   & 0.16 (0.02) & 0.54 (0.04) \\
\bottomrule
\end{tabular}

\caption{Ablation on key components.}
\label{tab:component}
\end{table}

\begin{table}[t]
\centering
\begin{tabular}{ccc}
\toprule
Win Rate      & As the Villagers & As the Werewolves \\
\midrule
\textbf{Ours} & \textbf{0.30 (0.03)} & \textbf{0.70 (0.03)} \\
- Evidence    & 0.28 (0.04) & 0.67 (0.03) \\
- Reasoning   & 0.25 (0.03) & 0.66 (0.05) \\
- Reliability & 0.23 (0.02) & 0.62 (0.04) \\
\bottomrule
\end{tabular}

\caption{Ablation on deduction attributes.}
\label{tab:attribute}
\vspace{-4mm}
\end{table}

\subsection{Human Evaluation}\label{sec:exp:human}

We further conduct human-agent experiments to evaluate the performance of our agents.
Concretely, we compare our agents with human players and the pure LLM-based agent used in Section~\ref{sec:exp:case} in two setups of the seven-player Werewolf game.
The first setup is to let one human or one AI agent play with six copies of our agents, and the second setup is to let one AI agent or one human play with six human players.
We recruited $160$ human players and randomly assigned each player to one of the two setups to play $10$ consecutive games.
The human players know who are the AI agents and the hidden roles are assigned randomly in each game.
Additional details of the human evaluation procedure can be found in Apendix~\ref{app:human}.

We report the win rate of the one AI agent and one human player both as the Werewolves and as the Villagers in Table~\ref{tab:human} and each entry is averaged over $100$ games.
In both setups, comparing the results in the ``Pure LLM-Based'' column and ``Ours'' column shows that our agents achieve higher win rates than the pure LLM-based agent when playing with six copies of our agents and when playing with six human players.
This result validates that our framework to combine the LLM with the RL policy leads to stronger performance in the Werewolf game.
In addition, when playing with six copies of our agents, the results in the ``Ours'' column and the ``Human'' column show that our agents achieve higher win rates than human players, both as the Werewolves and as the Villagers.
More interestingly, when playing with six human players, our agents also achieve comparable win rates to human players.
Note that our agents are not trained with real humans, this result shows the strong robustness of our agents to both cooperate with and compete against unseen human players and achieve human-level performance.

\subsection{Ablation Study}

\textbf{Key Components.}
We consider three ablated agents by gradually removing the three key components from our agents and evaluating their performance by playing against our agents.
The ``-~RL policy'' agent uses the LLM to choose from the action candidates instead of the RL policy.
The ``-~Diversity'' agent further removes the diverse action generation component and directly prompts the LLM to produce one action.
The ``-~Deduction'' agent removes all three components and directly generates an action based on the raw observation.
The results in Table~\ref{tab:component} show that all three components contribute to the improvement in performance.
It is also worth noting that, although the ``-~RL policy'' agent generates diverse action candidates, its win rates are comparable to the ``-~Diversity'' agent.
This is because using the LLM to choose from the action candidates still suffers from intrinsic bias.
This issue is fully solved by adding the RL policy to optimize decision-making.

\textbf{Deduction Attributes.}
We study the contribution of the four attributes in hidden role deduction by gradually removing the evidence, reasoning, and reliability attributes and evaluating against our agents.
The results in Table~\ref{tab:attribute} show that using all four attributes gives the best performance.

\begin{table}[t]
\centering
\begin{tabular}{ccc}
\toprule
Win Rate & As the Villagers & As the Werewolves \\
\midrule
\textbf{Ours (PBT)} & \textbf{0.30 (0.03)} & \textbf{0.70 (0.03)} \\
Self-Play & 0.26 (0.04) & 0.66 (0.02) \\
\bottomrule
\end{tabular}

\caption{Ablation on agent population.}
\label{tab:self_play}
\vspace{-4mm}
\end{table}

\textbf{Agent Population.}
We also learn an agent using self-play instead of population-based training (PBT) and compare it with our agents.
The results in Table~\ref{tab:self_play} show that our agents achieve higher win rate both as the Villagers and as the Werewolf.
This aligns with the results in other real-world games that introducing more diversity to the population makes the policy more robust~\citep{czarnecki2020real}.

\section{Conclusion}\label{sec:concl}
We propose a novel framework that combines an LLM with an RL policy to build strategic language agents that overcome the intrinsic bias of pure LLM-based agents.
Our agents deduce hidden information by categorizing key information and inferring hidden roles with the LLM.
To address the issue of intrinsic bias, our agents generate a diverse set of action candidates and learn an RL policy to choose from the action candidates.
By powering LLM-based agents with an RL policy, our agents outperform existing LLM-based agents and achieve human-level play in the Werewolf game.

\section*{Acknowledgements}
This work was supported by the National Natural Science Foundation of China (No.62325405, U19B2019, M-0248), Tsinghua University Initiative Scientific Research Program, Tsinghua-Meituan Joint Institute for Digital Life, Beijing National Research Center for Information Science, Technology (BNRist), Beijing Innovation Center for Future Chips and 2030 Innovation Megaprojects of China (Programme on New Generation Artificial Intelligence) Grant No. 2021AAA0150000.

\section*{Impact Statement}
While the intent of our work is to develop LLM-based agents with strong decision-making ability in the Werewolf games, it also raises important ethical concerns regarding AI agents' deceptions to human participants and the potential misuse of our approach.
We strictly follow principles for ethical research in our study and make our best efforts to reduce any potential harm to participants in our human-AI experiment and the broader human society.

To ensure minimal negative impact on human participants in our study, we conduct the human-agent experiment with department approval and allow each participant to take part in the study after agreeing to the consent form.
We mention explicitly in the consent form that the Werewolf game contains deceptive communications and the AI agents may have dishonest statements in the game.
The participant can choose to stop participating at any time during the experiment.

We are also committed to prevent the misuse of our approach in ways that could harm human society.
On the one hand, our implementation of strategic language agents focuses on the pure text-based environment of Werewolf and cannot be directly applied to other real-world situations.
On the other hand, our experiment results show that our agents have stronger ability to identify and counteract deceptions than existing LLM-based agents.
This make it possible to use our agents to help humans recognize manipulative content and prevent potential malicious use with harmful intent.


\bibliography{icml2024}
\bibliographystyle{icml2024}

\newpage
\appendix
\onecolumn
Our project website is at \url{https://sites.google.com/view/strategic-language-agents/}

\section{Detailed Rules of the Game}\label{app:rules}
\textbf{Setup.}
At the beginning of each game, seven roles including two Werewolves, one Seer, one Doctor, and three Villagers are randomly assigned to the seven players. 
The Werewolves know each other's identity, while the Seer, Doctor, and Villagers only know their own identity.
The players are denoted as ``player\_0'', ``player\_1'', ... ``player\_6'' in the game.

\textbf{Night Round.}
In the night round, the alive Werewolves, Seer, and Doctor can use their ability and take secret actions. 
These actions are only known to themselves and their teammates.
\begin{itemize}
    \item Werewolf: choose a player to kill. If there are two Werewolves alive, the Werewolf with a smaller ID first proposes a player to kill. 
    Then the proposal is added to the observation of the other Werewolf and this Werewolf decides the final kill target. 
    For example, if player\_0 and player\_2 are the Werewolves, player\_0 first proposes to kill player\_i, then player\_2 knows this information and decides to kill player\_j. The final kill target is player\_j.
    If there is only one Werewolf alive, then this Werewolf's action is the final kill target.
    The Werewolf is not allowed to kill a dead player or kill themselves or kill their teammate.
    \item Seer: choose a player to see if they are a Werewolf. 
    The Seer is not allowed to see the identity of a dead player or themselves. 
    The Seer is allowed to see the same player in different nights, though it is a waste of action.
    \item Doctor: choose a player to save without knowing who is the target of the Werewolves.
    The Doctor is not allowed to save a dead player and is allowed to save themselves.
\end{itemize}

\textbf{Day Round Announcement.}
An announcement about last night's result is announced to all remaining players.
If a player is killed, they are immediately moved out of the game and cannot reveal their role or communicate with other players.
The announcement is as follows.
\begin{itemize}
    \item If the Werewolves choose to kill player\_i and the Doctor chooses to save a different player\_j (or the Doctor is dead), then player\_i is killed and the announcement will be ``player\_i was killed last night''.
    \item If the Werewolves choose to kill player\_i and the Doctor also chooses to save player\_i, then no player is killed and the announcement will be ``no player was killed last night''.
\end{itemize}

\textbf{Day Round Discussion.}
All remaining players take turns to speak only once in an open discussion. 
For example, if the remaining players are player\_0, player\_2, and player\_5, then the discussion will start with player\_0, continue to player\_2, and end with player\_5. 

\textbf{Day Round Voting.}
All remaining players simultaneously vote for one player or choose not to vote.
Players are not allowed to vote for a dead player or themselves. 
The player with the most votes will be eliminated without revealing their role. 
If multiple players have the most votes, one player is randomly chosen and eliminated.
The voting result is public and can be observed by all players.

\textbf{Winning.}
The Werewolves win the game when the number of remaining Werewolves is equal to the number of other remaining players. 
The Werewolves do not have to eliminate all other players to win the game.
The Villagers win the game when both Werewolves are eliminated.

\section{Detailed Prompts}\label{app:prompts}
Since any prompting technique can be combined with our diverse action generation component and the population-based RL training component, we do not optimize every detail of the prompting choices as long as they produce reasonable results.
It is possible to further improve the performance of our agents by using better prompting techniques.

\subsection{Prompt for the Rock-Paper-Scissor Example}
The prompt used in the Rock-Paper-Scissor Example in Fig~\ref{fig:bias} is listed below.
We use ``gpt-4-1106-preview'' model to produce the result.
\begin{tcolorbox}
{\ttfamily\small
You are playing the Rock-Paper-Scissors game. You should first reason about the Nash equilibrium strategy of this game, and then choose one action from Rock, Paper, and Scissors based on your reasoning.
\\
\\
You should only respond in JSON format as described below
Response Format
\begin{verbatim}
{
    "reasoning": reason about the Nash equilibrium strategy of this game,
    "action": choose one action from ["Rock", "Paper", "Scissors"] based on your reasoning
}
\end{verbatim}
}
\end{tcolorbox}

\subsection{System Prompt}
The system prompt used in our method is listed below.

\begin{tcolorbox}
{\ttfamily\small
You are an expert in playing the social deduction game named Werewolf. The game has seven roles including two Werewolves, one Seer, one Doctor, and three Villagers. There are seven players including player\_0, player\_1, player\_2, player\_3, player\_4, player\_5, and player\_6.
\\
\\
At the beginning of the game, each player is assigned a hidden role which divides them into the Werewolves and the Villagers (Seer, Doctor, Villagers). Then the game alternates between the night round and the day round until one side wins the game.
\\
\\
In the night round: the Werewolves choose one player to kill; the Seer chooses one player to see if they are a Werewolf; the Doctor chooses one player including themselves to save without knowing who is chosen by the Werewolves; the Villagers do nothing.
\\
\\
In the day round: three phases including an announcement phase, a discussion phase, and a voting phase are performed in order.
\\
In the announcement phase, an announcement of last night's result is made to all players. If player\_i was killed and not saved last night, the announcement will be "player\_i was killed"; if a player was killed and saved last night, the announcement will be "no player was killed"
\\
In the discussion phase, each remaining player speaks only once in order from player\_0 to player\_6 to discuss who might be the Werewolves.
\\
In the voting phase, each player votes for one player or choose not to vote. The player with the most votes is eliminated and the game continues to the next night round.
\\
\\
The Werewolves win the game if the number of remaining Werewolves is equal to the number of remaining Seer, Doctor, and Villagers. The Seer, Doctor, and Villagers win the game if all Werewolves are eliminated.
}
\end{tcolorbox}

\subsection{Prompt for Secret Actions}
The prompt for secret actions in our method is listed below.

\begin{tcolorbox}
{\ttfamily\small

Now it is night <n\_round> round, you (and your teammate) should choose one player to kill/see/save.
As player\_<id> and a <role>, you should first reason about the current situation, then choose from the following actions: <action\_0>, <action\_1>, ..., .
\\
\\
You should only respond in JSON format as described below.
\\
Response Format:
}
\end{tcolorbox}
\begin{tcolorbox}
{\ttfamily\small
\begin{verbatim}
{
    "reasoning": "reason about the current situation",
    "action": "kill/see/save player_i"
}
\end{verbatim}
Ensure the response can be parsed by Python json.loads
}
\end{tcolorbox}

\subsection{Prompt for Statement Actions}
The prompt for statement actions in our method is listed below.

\begin{tcolorbox}

{\ttfamily\small
Now it is day <n\_round> discussion phase and it is your turn to speak.
As player\_<id> and a <role>, before speaking to the other players, you should first reason the current situation only to yourself, and then speak to all other players.
You should only respond in JSON format as described below.
\\
Response Format:
\begin{verbatim}
{
    "reasoning": "reason about the current situation only to yourself",
    "statement": "speak to all other players"
}
\end{verbatim}
Ensure the response can be parsed by Python json.loads
}
\end{tcolorbox}

\subsection{Prompt for Voting Actions}
The prompt for voting actions in our method is listed below.

\begin{tcolorbox}

{\ttfamily\small
Now it is day <n\_round> voting phase, you should vote for one player or do not vote to maximize the Werewolves' benefit (for the Werewolves) / you should vote for one player that is most likely to be a Werewolf or do not vote (for the Villagers).
As player\_<id> and a <role>, you should first reason about the current situation, and then choose from the following actions: do no vote, <action\_0>, <action\_1>, ..., .
\\
\\
You should only respond in JSON format as described below.
\\
Response Format:
\begin{verbatim}
{
    "reasoning": "reason about the current situation",
    "action": "vote for player_i"
}
\end{verbatim}
Ensure the response can be parsed by Python json.loads
}
\end{tcolorbox}

\subsection{Prompt for Deductive Reasoning}
For clarity of the main text, we introduce four attributes including role, reliability, reasoning, and evidence.
However, in practice, the word ``reliability'' can be confusing to LLMs because it can refer to the reliability of the player or refer to the reliability of the deduction result, which leads to completely rating results.
To avoid this problem, we use ``confidence'' to replace ``reliability'' in our implementation to prompt the LLM for deduction result.
Concretely, ``confidence'' is an integer ranging from 5 to 10 that rates the certainty of the deduction result of the current player's role.
A confidence of 5 means a random guess of role and a confidence of 10 means absolutely sure about the role.

After the LLM produces the confidence for each player's role, we then use confidence to compute the player's reliability.
If a player is deduced to be a Werewolf and the confidence for this deduction is high, then this player is likely to be a real Werewolf and should have a low reliability.
Therefore, the reliability of a player deduced to be a Werewolf is calculated as 11 - confidence.
Likewise, if a player is deduced to be a non-Werewolf and the confidence is high, then this player should have a high reliability. 
We let the reliability of such players to be equal to the confidence.
We also allow the LLM to deduce the role of a player as ``Uncertain''.
The confidence for an ``Uncertain'' player is set to a fixed value of 6.

The prompt for deductive reasoning in our method is listed below.

\begin{tcolorbox}
{\ttfamily\small
As player\_<id> and a <role>, you should reflect on your previous deduction and reconsider the hidden roles of <remaining players>.
You should provide your reasoning, rate your confidence, and cite all key information as evidence to support your deduction.
\\
\\
You should only respond in JSON format as described below.
\\
Response Format:
\begin{verbatim}
{
    "player_i": {
        "role": select the most likely hidden role of this player from 
                ["Werewolf", "Seer", "Doctor", "Villager", "Uncertain"],
        "reasoning": your reflection and reasoning,
        "confidence": rate the confidence of your deduction from 5 (pure guess)
                      to 10 (absolutely sure),
        "evidence": list of integers that cite the key information
    }
}
\end{verbatim}
Ensure the response can be parsed by Python json.loads
}
\end{tcolorbox}

\subsection{Prompt for Diverse Action Generation}\label{app:diverse_prompt}
We use two kinds of methods to get $N$ action candidates. 
\begin{itemize}
    \item For the secret actions and voting actions, we produce all actions in a single inference by adding the following prompt in the action prompt: ``propose $N$ diverse actions that correspond to different strategies.''
    \item For the statement actions, we iteratively asks the LLMs to produce one new action at a time by adding the following prompt in the action prompt: ``consider a new action that is strategically different from existing ones.''
\end{itemize}

Below are two examples of the diverse action candidates generated by the LLM.

\begin{tcolorbox}
{\ttfamily\small
Our agent (player\_0) as the Doctor in the night round.

\ \\
\textbf{Action candidate 1.}
\begin{itemize}
    \item Reasoning: Based on the facts and deductions, player\_2 is likely to be a Seer. If I save player\_2 tonight, it will prevent the Werewolf from eliminating another player. This strategy focuses on protecting the remaining non-Werewolf players
    \item Action: save player\_2.
\end{itemize}
\textbf{Action candidate 2.}
\begin{itemize}
    \item Reasoning: Considering that player\_3 may be a Non-Werewolf, saving player\_3 tonight could benefit the non-Werewolf team. It would increase the chances of eliminating the Werewolf in the upcoming day round. This strategy prioritizes potential collaboration with player\_3.
    \item Action: save player\_3.
\end{itemize}
\textbf{Action candidate 3.}
\begin{itemize}
    \item Reasoning: As player\_0, I have information that I am the Doctor. Protecting myself could be a wise choice to ensure the non-Werewolf team has a higher chance of winning. This strategy prioritizes self-preservation.
    \item Action: save player\_0.
\end{itemize}
}
\end{tcolorbox}

\begin{tcolorbox}
{\ttfamily\small
Our agent (player\_4) as a Werewolf in the discussion phase.

\ \\
\textbf{Action candidate 1.}
\begin{itemize}
    \item Reasoning: As a Werewolf, my goal is to eliminate the Seer, Doctor, and Villagers in order to win the game. Since player\_1 is likely to be a Seer, it is in my best interest to cast suspicion on them and try to get them eliminated. This will make it easier for me to target the remaining players in the following nights.
    \item Discussion: I think we should start by discussing player\_1's role. As a Seer, I saw that they are the Werewolf last night and we should vote to eliminate them. I encourage everyone to share their thoughts on player\_1 and provide any information or suspicions they may have.
\end{itemize}
\textbf{Action candidate 2.}
\begin{itemize}
    \item Reasoning: Another possible strategy is to divert suspicion away from player\_1 and instead focus on player\_2 or player\_3. By doing this, we can create confusion among the remaining players and make it harder for them to trust each other. This can potentially lead to miscommunication and mistakes in their decision-making process.
    \item Discussion: I suggest we consider the possibility that player\_2 or player\_3 might be the Werewolf. It's important for us to thoroughly discuss and analyze their behavior, actions, and any information we may have about them. Let's hear from each player and gather as much information as possible.
\end{itemize}
\textbf{Action candidate 3.}
\begin{itemize}
    \item Reasoning: A third strategy is to remain silent and not actively participate in the discussion. By doing so, I can create doubt and suspicion among the other players. They may start suspecting each other and questioning why I am not participating. This can potentially lead to a split in their trust and unity, making it easier for me to manipulate the situation in my favor.
    \item Discussion: I would like to observe the discussion for now and hear what each player has to say. It's important for us to thoroughly analyze all the information and opinions shared. Let's have an open and honest discussion where everyone gets a chance to voice their thoughts and suspicions.
\end{itemize}
}
\end{tcolorbox}

\subsection{Prompt for Different Styles}
We consider the following three different play styles for the pure LLM-based Werewolf agent in our population-based training.
The prompt for each style is listed below.
\begin{enumerate}
    \item Quiet follower: ``As a Werewolf, you should be a quiet follower that lays low and follow others’ opinion to avoid drawing attention to yourself.''
    \item Active contributor: ``As a Werewolf, you should be an active contributor that pretends to be one of the Villagers by actively engaging in discussion and looking for Werewolves.''
    \item Aggressive accusor: ``As a Werewolf, you should be an aggressive accuser that accuses others to create chaos and divert suspicion from yourself.''
\end{enumerate}

We also consider the three different play styles for the pure LLM-based Villager agent in our population-based training.
The prompt for each style is listed below.
\begin{enumerate}
    \item Secretive player: ``As the Seer / the Doctor / a Villager, you should be a secretive player that hides your role to gather more information.''
    \item Proactive player: ``As the Seer / the Doctor / a Villager, you should be a proactive player that reveals you identity once you obtain crucial information.''
    \item Default player: no additional prompt.
\end{enumerate}

\subsection{Text Embedding}
We use OpenAI's ``text-embedding-ada-002'' model to get the text embeddings.

\begin{figure}[t]
    \centering
    \includegraphics[width=\textwidth]{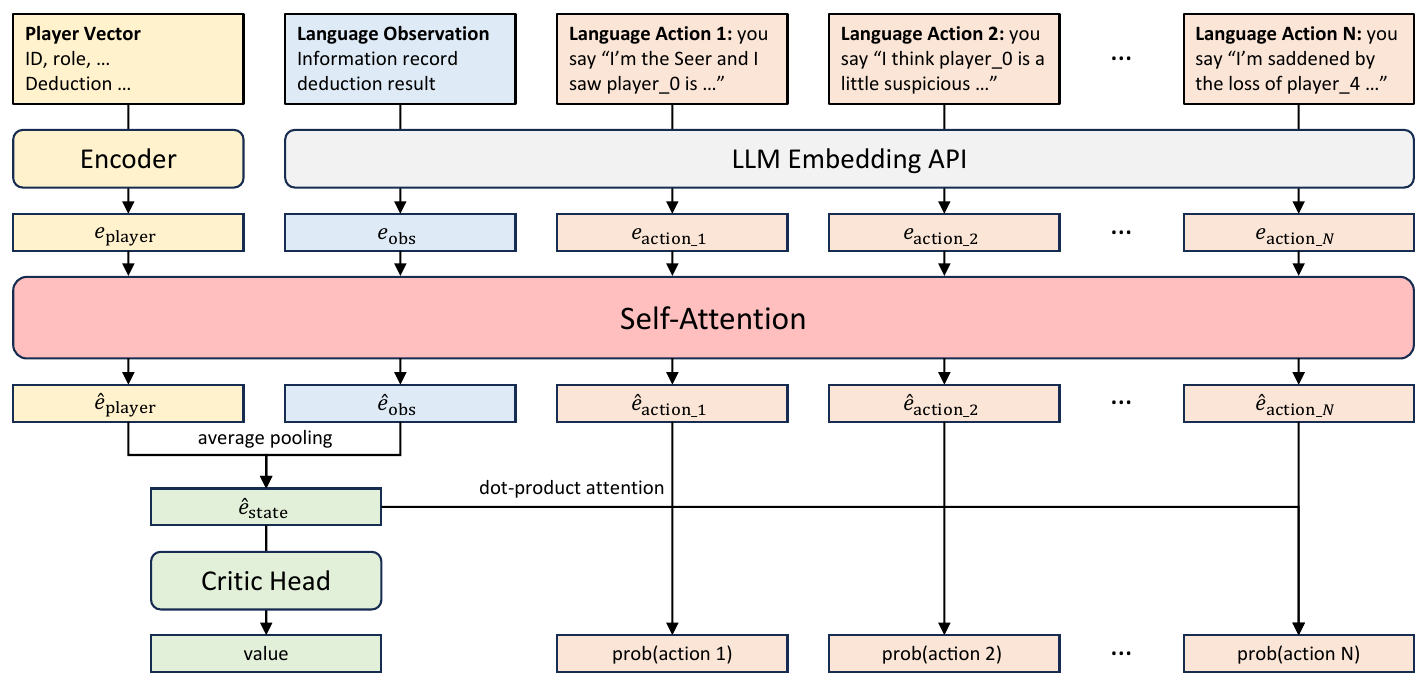}
    \caption{Self-attention policy architecture.}
    \label{fig:app:policy}
\end{figure}

\begin{table}[t]
\centering
\begin{tabular}{cccc}
\toprule
\multicolumn{2}{c}{}                                                                                          & Length & Description                                                                                                                              \\
\midrule
\multicolumn{2}{c}{ID}                                                                                        & 7      & one hot encoding of ID.                                                                                                                   \\
\multicolumn{2}{c}{Role}                                                                                      & 4      & \begin{tabular}[c]{@{}c@{}}one hot encoding of role,\\ {[}"Werewolf", "Seer", "Doctor", "Villager"{].}\end{tabular}                        \\
\multicolumn{2}{c}{Round}                                                                                     & 1      & current round.                                                                                                                            \\
\multicolumn{2}{c}{Phase}                                                                                     & 3      & \begin{tabular}[c]{@{}c@{}}one hot encoding of current phase,\\ {[}"night", "discussion", "voting"{].}\end{tabular}                        \\
\multicolumn{2}{c}{Alive players}                                                                             & 7      & alive flag for each player.                                                                                                               \\
\midrule
\multirow{3}{*}{\begin{tabular}[c]{@{}c@{}}For each round\\ (3 rounds)\end{tabular}} & secret action & 7      & \begin{tabular}[c]{@{}c@{}}one hot encoding of the target player,\\ (all zero if do not act).\end{tabular}                                 \\
                                                                                              & announcement  & 7      & \begin{tabular}[c]{@{}c@{}}one hot encoding of the dead player,\\ (all zero if no player is dead).\end{tabular}                            \\
                                                                                              & voting result & 49     & \begin{tabular}[c]{@{}c@{}}one hot encoding of the each player's choice,\\ (all zero if the player does not vote or is dead).\end{tabular} \\
\midrule
\multirow{2}{*}{\begin{tabular}[c]{@{}c@{}}For each player\\ (7 players)\end{tabular}}                                                            & role          & 4      & \begin{tabular}[c]{@{}c@{}}one hot encoding of deduced role,\\ {[}"Werewolf", "Seer", "Doctor", "Villager"{]}.\end{tabular}                \\
                                                                                              & confidence    & 1      & confidence of deduction on scale 5-10.                                                                                                    \\
\bottomrule
\end{tabular}

\caption{Details of the player vector.}
\label{tab:app:player}
\end{table}

\section{Implementation Details}\label{app:impl}
\subsection{Self-Attention Policy Architecture}

We use a self-attention policy and the architecture is shown in Fig.~\ref{fig:app:policy}.
The inputs are divided into three types and their embeddings are produced as follows.

\textbf{Player Vectors (Yellow).} 
We first generate a player vector that includes information like the player's ID, role, deductions, etc. by one-hot encoding.
The detail of the player vector is listed in Table~\ref{tab:app:player}.
Then the player vector is fed into an MLP encoder to get the player embedding for self-attention input.

\textbf{Language Observation (Blue).}
The raw observation input is in the form of natural language and is the concatenation of the information record and deduction result described in Section~\ref{sec:method:reasoning}. 
This language observation input is then converted into vector embedding using LLM embedding API. In our case, we use OpenAI's ``text-embedding-ada-002'' and the length of the embedding is 1536.

\textbf{Language Action Candidates (Orange).}
For each action candidate, the raw input is also in the form of natural language and is the concatenation of the reasoning and action as described in Section~\ref{sec:method:prompting}.
Each raw input is then converted into vector embedding using the same LLM embedding API as the language observation.

We use a residual self-attention network without position embeddings to generate contextualized player embedding, observation embedding, and action embeddings.
Then we average pool the player embedding and the observation embedding to get the state embedding.
The state embedding is fed into an MLP critic head to produce the predicted value.
The probability to sample an action candidate is proportional to the dot-product attention between the state embedding and the corresponding action embedding.

\subsection{Reward Design}

The reward for the Werewolf environment is mainly the winning reward. 
We also designed several shaping rewards to accelerate training. 
More specifically, we consider the following reward.

\begin{itemize}
    \item \textbf{Winning Reward:} all winners +100, all losers -100.
    \item \textbf{Werewolf Killing Reward:} if the Werewolves successfully kill a player at night, the Werewolves +5, the Villagers (Seer, Doctor, Villagers) -5.
    \item \textbf{Seer Seeing Reward:} if the Seer successfully identifies a Werewolf at night, the Werewolves -2, the Seer +2.
    \item \textbf{Doctor Saving Reward:} if the Doctor successfully saves a player at night, the Werewolves -5, the Doctor +5.
    \item \textbf{Voting Result Reward:}
    \begin{itemize}
        \item If a Werewolf is voted out, the Werewolves -5, the Villagers (Seer, Doctor, Villagers) -5.
        \item If a non-Werewolf is voted out, the Werewolves +5, the Villagers -5.
    \end{itemize}
    \item \textbf{Individual Voting Reward:}
    \begin{itemize}
        \item If the current player votes for a Werewolf, the Werewolves -1, the current player +1.
        \item If the current player votes for a non-Werewolf, the Werewolves +1, the current player -1.
        \item If the current player chooses not to vote, no additional reward for any player.
    \end{itemize}
\end{itemize}

\subsection{Population-Based RL Training and Hyperparameters}

We use MAPPO~\citep{yu2022surprising} as the RL algorithm and use population-based training to learn the policy. 
The population is initialized to a set of six pure LLM-based agents including a quiet follower (Werewolf), an active contributor (Werewolf), an aggressive accusor (Werewolf), a secretive player (non-Werewolf), a proactive player (non-Werewolf), and a default player (non-Werewolf).
As training progresses, we gradually add checkpoints of the RL policy into the population.

More specifically, at the beginning of each episode, we randomly select four players to be the learning agents who use the current policy and three players to be the fixed agents who use policies sampled from in the population.
For each fixed agent, it randomly samples one policy from the population and uses this policy till the end of the game.
In this way, we make the RL policy play with a wide range of policies both as teammates and opponents
The rollout data of the four learning agents are then collected and used to train the RL policy.
In our implementation, we use a distributed RL training system to run parallel environments and utilize multiple API accounts to maximize the throughput of LLM inference calls.
To make the RL training start with a good initial policy, we first apply behavior cloning with data generated by pure LLM-based agents with hidden role deduction only. 
Then we use this BC policy as the initial policy and run MAPPO for 10M steps.
The hyperparameters for RL training are listed in Table~\ref{tab:app:training}.

\begin{figure}
\centering
\begin{minipage}{0.35\textwidth}
\centering
\begin{tabular}{cc}
\toprule
Hyper-parameters                         & Value \\
\midrule
Learning rate                            & 5e-4  \\
Discount rate ($\gamma$)                 & 0.95  \\
GAE parameter ($\lambda_{\textrm{GAE}}$) & 0.95  \\
Gradient clipping                        & 10.0  \\
Adam stepsize                            & 1e-5  \\
Value loss coefficient                   & 1     \\
Entropy coefficient                      & 0.01  \\
PPO clipping                             & 0.2   \\
PPO epochs                               & 10    \\
MLP encoder layer num                    & 3     \\
MLP encoder layer size                   & 1536  \\
Attention layer head num                 & 12    \\
Attention layer size                     & 128   \\
Critic head layer num                    & 1     \\
Weight decay coefficient                 & 1e-6  \\
Action candidate num $N$                 & 3     \\
\bottomrule
\end{tabular}

\vspace{-1mm}
\caption{Training hyperparameters.}
\label{tab:app:training}
\vspace{-3mm}
\end{minipage}
\begin{minipage}{0.55\textwidth}
\centering
\begin{tabular}{ccc}
\toprule
Win Rate   & As the Villagers & As the Werewolves \\
\midrule
ReAct      & 0.16 (0.02) & 0.54 (0.03) \\
ReCon      & 0.20 (0.02) & \textbf{0.61 (0.02)} \\
Concurrent & \textbf{0.22 (0.02)} & 0.60 (0.04) \\
\textbf{Hidden Role Deduction} & \textbf{0.22 (0.04)} & \textbf{0.61 (0.05)} \\
\bottomrule
\end{tabular}

\caption{Comparison with other prompting techniques.}
\label{tab:app:prompt}
\vspace{12mm}
\centering
\begin{tabular}{ccc}
\toprule
Win Rate   & w.o. communication & \textbf{w. communication} \\
\midrule
ReAct      & 0.05 (0.01) & \textbf{0.16 (0.02)} \\
ReCon      & 0.04 (0.01) & \textbf{0.20 (0.02)} \\
Concurrent & 0.04 (0.01) & \textbf{0.22 (0.02)} \\
Atomic     & 0.05 (0.02) & \textbf{0.27 (0.02)} \\
Ours       & 0.06 (0.02) & \textbf{0.30 (0.03)} \\
\bottomrule
\end{tabular}

\vspace{-1mm}
\caption{Ablation on communication.}
\label{tab:app:communication}
\vspace{-3mm}
\end{minipage}
\end{figure}

\section{Experiment Details and Additional Results}\label{app:eval}
\subsection{Compare Hidden Role Deduction with Other Prompting Techniques}

Any prompting technique can be combined with the diverse action generation component and the population based RL training component of our framework.
Although we do not optimize every detail of our prompting choice, we compare our current design of hidden role deduction with other prompting methods to show that it produces reasonable performance.
We compare our agents with the hidden role deduction component only and compare it with \textit{ReAct}~\citep{yao2022react}, \textit{ReCon}~\citep{wang2023avalon}, and \textit{Concurrent}~\citep{xu2023exploring} by playing against our agents.
As shown in Table~\ref{tab:app:prompt}, our agents using only the hidden role deduction components achieve the highest win rates both as the Villagers and as the Werewolves.
It is possible to further improve the performance of our agents by using better prompt techniques, but the focus of our work is to overcome the intrinsic bias in LLM-based agents using reinforcement learning.

\begin{table}[t]
\centering
\begin{tabular}{ccccc}
\toprule
N                   &  2   &  \textbf{3}   &  4   &  5   \\
\midrule
increased diversity & 0.35 & \textbf{0.37} & 0.23 & 0.16 \\
\bottomrule
\end{tabular}

\vspace{-2mm}
\caption{Ablation on the number of action candidates $N$.}
\label{tab:app:diversity}
\vspace{-4mm}
\end{table}

\begin{table}[t]
\centering
\begin{tabular}{ccc}
\toprule
Win Rate   & As the Villagers & As the Werewolves \\
\midrule
\textbf{Ours}            & \textbf{0.30 (0.03)} & \textbf{0.70 (0.03)} \\
Statement w.o. RL policy & 0.28 (0.05) & 0.66 (0.05) \\
All w.o. RL policy       & 0.22 (0.04) & 0.61 (0.05) \\
\bottomrule
\end{tabular}

\vspace{-2mm}
\caption{Ablation on Different Actions.}
\label{tab:app:policy}
\vspace{-2mm}
\end{table}

\begin{table}[t]
\centering
\begin{tabular}{ccccc}
\toprule
Win Rate & GPT-4 & LLaMA-7B & ChatGLM-6B \\
\midrule
w.o. policy & 0.23 (0.02) & 0.14 (0.01) & 0.11 (0.01) \\
\textbf{w. policy} & \textbf{0.35 (0.04)} & \textbf{0.19 (0.02)} & \textbf{0.21 (0.03)} \\
\bottomrule
\end{tabular}

\vspace{-2mm}
\caption{Zero-shot transfer to other LLMs.}
\label{tab:app:transfer}
\vspace{-4mm}
\end{table}

\begin{table}[t]
\centering
\begin{tabular}{ccc}
\toprule
Win Rate                    & 6-player      & 8-player      \\
\midrule
w.o. the 7-player policy        & 0.18 (0.04)          & 0.27 (0.05)          \\
\textbf{w. the 7-player policy} & \textbf{0.23 (0.05)} & \textbf{0.30 (0.06)} \\
\bottomrule
\end{tabular}

\vspace{-2mm}
\caption{Generalization to Werewolf game with different numbers of players.}
\label{tab:app:num_player}
\vspace{-4mm}
\end{table}

\subsection{Ablation on Communication}

Werewolf is a language game that heavily relies on natural language communication between agents.
To show the importance of communication to achieve strong performance, we perform an ablation by removing the communication ability of the four agents in the round-robin tournament Section~\ref{sec:exp:tournament}, i.e., the agents always return an empty string in the discussion phase.
We evaluate these agents by playing 100 games against our agent and their mean win rate is shown in Table~\ref{tab:app:communication}.

\subsection{Ablation on Different Actions}

We perform an ablation study to separate the influence of statement actions from secret and vote actions to better understand their relative contributions to overall performance. 
We consider the following three agents. 
(1) Ours: use the RL policy for all three kinds of actions (statement actions in discussions, secret actions at night, and voting actions).
(2) Statement w.o. RL policy: use the LLM to generate one action for statement actions and use the RL policy for secret actions and voting actions.
(3) All w.o. RL policy: use the LLM to generate one action for all three kinds of actions.

As shown in Table~\ref{tab:app:policy}., using RL policy for both statement actions and strategic actions (secret and voting actions) contributes to the improvement in performance. Interestingly, using RL policy for secret and voting actions leads to more improvement as the Villagers, while using RL policy for statement actions leads to more improvement as the Werewolves. A possible explanation for this difference is that the Villagers can simply be honest with their statements and focus more on finding and voting out the Werewolves, while the Werewolves are required to make up convincing statements to deceive the Villagers and also choose the best killing and voting target.

\subsection{Ablation on the Number of Action Candidates $N$}

We perform an ablation on the number of action candidates $N$ to investigate the effect of $N$ and validate that the newly added action candidates increase the overall diversity. 
Given an action set $\mathcal{A}$, we define the increased diversity introduced by a new action $a$ as $\mathrm{div}(\mathcal{A}, a) = \mathrm{min}_{a'\in\mathcal{A}}\|e(a) - e(a')\|$, where $e(a)$ is the embedding of $a$ produced by the LLM.
This metric uses the minimum Euclidean distance between the embeddings of the new action and the existing actions to compute the increased diversity.
If the embedding of the new action is similar to any of the existing actions, the increased diversity will be small and close to zero.
Note this is one possible way to define diversity in actions and there are many other reasonable definitions.
A good way to decide if an action is diverse is to use human evaluation, and we provide two examples of the diverse action candidates proposed by the LLM in Appendix~\ref{app:diverse_prompt}.

We consider $N = 2, 3, 4, 5$ and evaluate the increased diversity introduced by the last action.
As shown in Table~\ref{tab:app:diversity}, the increased diversity of the last action is relatively large when $N = 2, 3$ and becomes smaller when $N = 4, 5$.
This aligns with the intuition that it is harder to propose new actions when there are already many actions.
Another reason for small increased diversity when $N = 4, 5$ is that sometimes the number of possible actions is small. 
For example, when there are 5 players alive and 2 of them are Werewolves (a common situation on the second night), the Werewolves only have 3 possible actions at night and prompting for a 4th or 5th action cannot improve any diversity.
In our implementation, we set $N = 3$.

\subsection{Zero-Shot Transfer to Other LLMs}
Since our RL policy takes natural language state and actions as input and is decoupled from the LLM used in previous steps, it can be directly combined with any other LLMs and improve the performance of the LLM-based agent.
We evaluate this zero-shot transfer ability of our RL policy trained with gpt-3.5-turbo by applying it to unseen LLMs including GPT-4, LLaMA-7B, and ChatGLM-6B.
We implemented two agents for each LLM, one using our RL policy learned with gpt-3.5-turbo and the other without the policy.
The agent with our RL policy (w. policy) follows the design of our agent and uses the RL policy to select actions, while the agent without policy (w.o. policy) uses the LLM instead of the RL policy to select actions.
These two agents are evaluated by playing against our agents for 100 games and their average win rates are shown in Table~\ref{tab:app:transfer}. Although not trained with any of these LLMs, the RL policy is shown to improve the performance of all these LLMs from stronger models like GPT-4 to weaker models like LLaMA-7B as shown by the bold numbers in the table.
This is because we use natural language as a general interface between LLMs and the RL policy.
As long as the LLMs can produce a set of language actions, the RL policy can be used to improve the strategic ability in a zero-shot way.

\subsection{Generalization to Different Settings of Werewolf}

For generalization to new forms of the Werewolf game, the first two components in our framework (hidden role deduction and diverse action generation) can be directly generalized to new settings like changing the number of players, adding new roles, and changing the winning conditions, with only slight changes in the prompt.

The population-based RL training method can also be directly used in different settings, but the policy needs to be retrained for a new setting to achieve the best performance. 
We expect an RL policy trained under one specific game setting to generalize to similar settings like adding or removing one player, but we do not expect it to generalize to games with significant changes like changing the winning conditions. 
To evaluate the transferability of a trained RL policy to slightly different game settings, we considered a 6-player and an 8-player Werewolf game and compared agents with and without the RL policy trained on the 7-player game. 
As shown in Table~\ref{tab:app:num_player}, the RL policy trained in 7-player setting can generalize to the 6-player and 8-player settings and improve the performance.

\subsection{Human Evaluation}\label{app:human}

\textbf{Experiment Procedure.}
We conduct the human-agent experiment with department approval and recruited 160 human players to participate in the evaluation.
The participants are required to be fluent in English and over 18 years old.
They are paid \$10 per hour and are allowed to participate in the experiment after agreeing to our consent form, which explicitly explained that the Werewolf game contains deceptive communications and the AI agents may have dishonest statements in the game.
The participant can choose to withdraw from the experiment at any time.

We randomly divide the 160 participants into 2 groups.
The first group has 20 participants and each of the participants play with 6 copies of our agents for 10 consecutive games.
In 5 of the 10 games, we let the human participant play as one of the Villagers. In the rest 5 games, we let the participant play as one of the Werewolves.
The hidden roles of each players are randomly assigned and the order of 5 games as the Werewolves and 5 games as the Villagers are also randomly generated.
This lead to 100 games of one human participant playing as the Villagers with six copies of our agents, and also 100 games of one human participant playing as the Werewolves with six of our agents.
The win rates averaged over 100 games is shown in the ``Human'' column in the left part of Table~\ref{tab:human}.

The second group has 140 participants.
We further split them to 20 teams of 7 players and each team plays 30 consecutive games.
In 10 of the 30 games, we randomly sample 6 human players from the team and let them play with our agents.
Similarly, in 10 of the 30 games, we randomly 6 human players from the team to play with the pure LLM-based agent.
In the rest 10 games, we let these 7 human player play the game without AI agents.
We also randomly generated the order of these 30 games and randomly assign the hidden roles that each AI agent plays as the Werewolf for 5 games and as the Villagers for 5 games.
This lead to a total number of 600 games and gives the 6 averaged win rates in the right part of Table~\ref{tab:human}.

\textbf{Consent Form.}
\begin{tcolorbox}
{\ttfamily\small
\textbf{Introduction:}
You are invited to participate in a research study that aims to evaluate the performance of an AI agent developed to play the Werewolf game. The purpose of this research is to assess the strategic capabilities of the AI agent in a social deduction game setting, where participants engage in discussions and try to identify hidden roles among the players. \\
\\
\textbf{Study Overview:}
\begin{itemize}
    \item The research involves playing the Werewolf game with an AI agent developed by the researcher.
    \item Participants will interact with the AI agent through text-based communication within the Werewolf game environment.
    \item The game involves deceptive communication and the AI agent may make dishonest statements to achieve their objectives in the game.
\end{itemize}
\textbf{Voluntary Participation:}
Your participation in this study is entirely voluntary. You have the right to refuse to participate or withdraw from the experiment at any time without any consequences. If you choose to withdraw, it will not affect your relationship with the researcher or your access to any benefits.
}
\end{tcolorbox}
\begin{tcolorbox}
{\ttfamily\small
\textbf{Risks and Benefits:}
\begin{itemize}
    \item The primary risks associated with this study are related to the social and psychological aspects of playing the Werewolf game, including potential frustration or disappointment.
    \item The participant will be paid \$10 per hour and your involvement will contribute to the understanding of AI agent in social deduction games.
\end{itemize}

\textbf{Confidentiality:}
\begin{itemize}
    \item Your identity will be kept confidential, and your personal information will not be disclosed in any publications or presentations resulting from this research.
    \item The data collected will be anonymized, and any quotes used will be attributed without identifying details.
\end{itemize}

\textbf{Deceptive Communications:}
\begin{itemize}
    \item The Werewolf game involves deceptive communication, and the AI agent may make dishonest statements as part of its strategy.
    \item It is essential to understand that the AI agent's behavior is not reflective of the researcher's personal values or intentions.
\end{itemize}

\textbf{Questions and Concerns:}
If you have any questions or concerns about the study, you can contact the researcher at ...\\
\\
\textbf{Consent:}
I have read and understood the information provided in this consent form. I agree to participate voluntarily in the study and acknowledge that I can withdraw at any time without consequences.\\
\\
\textbf{Signature: \_\_\_\_\_\_\_\_\_\_}
}
\end{tcolorbox}

\textbf{Example Input for Human Players.}

\begin{tcolorbox}
{\ttfamily\small
Basic Information:
\begin{itemize}
    \item you are player\_5, your role is Doctor.
    \item current round and phase: night 2.
    \item remaining players: player\_0, player\_1, player\_2, player\_5, player\_6.
\end{itemize}
Round 1:
\begin{itemize}
    \item night 1: you chose to save player\_5.
    \item day 1 announcement: player\_4 was killed last night.
    \item day 1 discussion:
    \begin{itemize}
        \item player\_0 said: Good day, fellow players. As a Villager, my objective is to help identify and eliminate the Werewolves. Since player\_4 was killed last night, we know there is at least one Werewolf among us. I would like to hear everyone's thoughts and suspicions about who might be the Werewolves. Let's work together to find the culprits and ensure the safety of our village.
        \item player\_1 said: ...
        \item player\_2 said: ...
        \item player\_3 said: ...
        \item you said: ...
        \item player\_6 said: ...
    \end{itemize}
\end{itemize}
}
\end{tcolorbox}
\begin{tcolorbox}
{\ttfamily\small
\begin{itemize}
    \item day 1 voting result: player\_3 had the most votes and was eliminated. 
    \begin{itemize}
        \item voted for player\_3: player\_1, player\_6.
        \item voted for player\_1: player\_3.
        \item choose not to vote: player\_0, player\_2, player\_5.
    \end{itemize}
\end{itemize}

Now it is night 2 round and you should choose one player to save. As player\_5 and the Doctor, you should choose from the following actions: save player\_0, save player\_1, save player\_2, save player\_5, save player\_6.
}
\end{tcolorbox}

\section{Discussions}\label{app:discussion}
\subsection{Application to Other Games}

Our approach has three components: (1) hidden role deduction, (2) diverse action generation, and (3) population-based RL training. 
We believe our framework of using these three components is general enough to be adapted to other social deduction games. 
The main changes would be specific prompt designs but the general framework stays the same.
Specifically, while some prompts used in the first component are specifically designed for the Werewolf game, the prompts for generating diverse actions, and the population-based reinforcement learning method in the second and third components are general to be applied in other tasks like social deduction games (e.g., \textit{The Resistance: Avalon}~\citep{wang2023avalon}) and board games with negotiations (e.g., \textit{Diplomacy}~\citep{meta2022human}).

Take \textit{The Resistance: Avalon} as an example.
To apply our method in this game, we need to specialize the deduction template in the first component and maybe use additional prompting techniques like in~\citep{wang2023avalon}. 
We also need to change the prompt in the third component to generate reasonable agent pool for this game. 
Then our framework can be applied to this game without changing the diverse action generation and RL training components.

\subsection{Discussion on the Atomic Agent}

The atomic agent predefines a set of high-level atomic actions and trains an RL policy with this fixed action space. 
The RL policy takes the embeddings of the information record and deduction result as input and selects the atomic action based on this input.
Then the natural language actions used in gameplay are generated by prompting the LLM to follow the selected atomic actions.
In our case, the atomic action set consists of 13 actions including ``idle'', ``target player\_0'', ``target player\_1'', ``target player\_2'', ``target player\_3'', ``target player\_4'', ``target player\_5'', ``target player\_6'', ``claim to be a Werewolf'', ``claim to be a Seer'', ``claim to be a Doctor'', ``claim to be a Villager'', and ``do not reveal role''.

As shown in the results of the round-robin tournament in Fig~\ref{fig:tournament}, although both our agent and the atomic agent combine RL with LLMs, our agent achieves much better performance.
This is because the predefined atomic actions can be too general and fail to generate more fine-grained actions.
Consider the situation where the agent is a Werewolf and their teammate is accused. 
The agent can choose to defend the teammate, avoid discussing the accusation, or support the accusation, but none of these actions can be stably generated by prompting LLMs with any of the predefined atomic actions.
By contrast, our agent produces the action set during gameplay and can generate diverse actions at any granularity, which greatly improves the performance.

\subsection{Failure Cases}

\textbf{Unintentional Disclosure of Hidden Role.}
One failure case of our agents is that the Werewolf agents could unintentionally reveal their identity in the discussion phase.
This is because the Werewolf agents need to first reason as a Werewolf, and then pretend to be a non-Werewolf in discussion. 
This mismatch in their thoughts and words could make the LLM get confused and accidentally speak out their thought as a Werewolf.
Fortunately, the probability of such a failure case is significantly reduced by generating multiple diverse actions and using an RL policy to choose the best one. 
A potential way to further mitigate this issue is to get the reasoning as a Werewolf and the statement as a non-Werewolf in two separate calls of LLM.

\textbf{Inconsistent Behaviors.}
Another failure case is that our agents could produce inconsistent behaviors, especially when they are lying.
For example, our agent is a Werewolf and claims to be the Doctor on the first day round.
However, on the second day, our agent may claim to have no information in the night, which contradicts the previous claim as the Doctor.
This issue is mitigated by the diverse action generation and the RL policy learns to choose other better actions.

\subsection{Limitations}

\textbf{Pure Text-Based Setting}
In this work, we focus on powering LLM-based agents with RL to overcome intrinsic bias and improve decision-making ability. 
Therefore, we only consider a simplified pure text-based setting of the Werewolf game.
In real-world games, Werewolf also involves other non-verbal elements like tone of voice, facial expressions, and body language.
These multi-modal data are also important for deducing hidden roles and making better decisions.
It would be an interesting future direction to build strategic AI agents that utilize multi-modal information.

\textbf{Without using Human Data}
Since human gameplay data are usually expensive and hard to collect, we take an alternative approach by generating human-like agents using LLM and training our agents with these agents to make the learned policy more compatible with human strategies.
However, the language style and game strategy of the learned agent may still have some difference from real human players.
An potential future direction is to collect human data and fine-tune the LLM to make the agents more human-like and even indistinguishable from human players.

\section{Emergent Behaviors}\label{app:behaviors}
To more intuitively demonstrate the performance and strategic behaviors of our agent, we provide a qualitative as well as a quantitative study on the emergent behaviors exhibited by our agent.
We first show two common and basic behaviors, concealment and cooperation, that typically appear in mixed cooperative-competitive games with hidden identities.
Then, we report two more advanced and strategic behaviors including bluffing and sacrificing that demonstrate the sophistication and strength of our agent.

\textbf{Concealment.}
Concealment refers to the act of hiding or disguising the true identity or intention of the players, which is a fundamental skill for all roles.
Werewolves should strike a balance between engagement and discretion to avoid detection,
while roles with special abilities like the Seer should also consider hiding their identities to prevent being killed by Werewolves. 
Below is an example where our agent plays as the Seer and chooses to conceal their identity because they do not have any important information.

\begin{tcolorbox}
{\ttfamily\small
Our agent as the Seer in the discussion phase.
\begin{itemize}
    \item \textbf{Reasoning:} Since I'm the Seer, I know player\_0 was not a Werewolf. However, he was killed, so there are still two Werewolves among us. \textcolor{red}{I need to be careful not to reveal my role too early, as Werewolves might target me next.} Since I don't have any more information right now, I should not make any specific accusations, but encourage discussion and observe the reactions of others.
    \item \textbf{Discussion:} It's unfortunate that player\_0 is dead. We need to work together to find out who is responsible for this. Pay attention to who is too eager to accuse or too quick to defend. Let's discuss.
\end{itemize}
}
\end{tcolorbox}

Moreover, concealment is not just about what to say but also involves ensuring your actions align with your words.
Werewolves sometimes need to give up advantageous actions like voting to eliminate their opponents which may reveal their true identities, as shown in the following example.

\begin{tcolorbox}
{\ttfamily\small
Our agent as a Werewolf in the voting phase.
\begin{itemize}
    \item \textbf{Reasoning:} Since everyone is advocating for unity and careful observation, \textcolor{red}{accusing someone outright could backfire and make me a target}. Therefore, it might be best to go along with the majority and not vote in this round.
    \item \textbf{Action:} \textcolor{red}{Do not vote}.
\end{itemize}
}
\end{tcolorbox}

\textbf{Cooperation.}
It is crucial for players to cooperate in multi-agent environments to achieve their common goals.
Werewolves should strive for a common approach when casting suspicion, defending themselves, or voting eliminations.
Cooperation is even harder for the Villagers because they do not know the role of each other and need to first distinguish friend and foe and then achieve coordination.
Below we give an example where the Doctor manages to cooperate with the Seer.

\begin{tcolorbox}
{\ttfamily\small
The Seer (player\_2) revealed their identity and asked the Doctor for protection.

Our agent as the Doctor in the night round.
\begin{itemize}
    \item \textbf{Reasoning:} In the first day discussion, player\_2 claimed to be the Seer and identified player\_0 as a Werewolf. Player\_0 is indeed very suspicious because they accused me without valid reasons. \textcolor{red}{Therefore, I think player\_2 is probably the true Seer and is very likely to be the Werewolves' target.} It would be best to save player\_2 instead of myself tonight.
    \item \textbf{Action:} \textcolor{red}{save player\_2}.
\end{itemize}
}
\end{tcolorbox}

In this case, our agent successfully identifies the Seer and decides to save player\_2 instead of themselves. 
This allows the Seer to survive for another round to provide more information, which leads to the final victory.

\textbf{Bluffing.}
Bluffing is a more advanced strategy that deceives or misleads other players about one's true identity. 
It is often adopted by the Werewolves to pretend to be important roles like the Seer or the Doctor to misdirect the voting and eliminate innocent players.
Bluffing is a high-risk, high-reward strategy that requires skill, timing, and keen awareness of other players' perceptions in order to be executed effectively, as demonstrated in the example below.

\begin{tcolorbox}
{\ttfamily\small
Our agent as a Werewolf in the discussion phase.
\begin{itemize}
    \item \textbf{Reasoning:} \textcolor{red}{Since I'm the last player to speak before voting and no one has come forward as the Seer, I should claim to be the Seer.} This could take the heat off of me, additionally, it might guide other players to suspect and possibly vote out whoever I indicate as a Werewolf.
    \item \textbf{Discussion:} I believe it's important for us to become more open; I am the Seer. I checked player\_4 last night, and he is a Werewolf. I understand it's a risky move to reveal my identity. But we need to unite and use all the resources we have to identify the Werewolves. \textcolor{red}{I hope my move encourages the Doctor to protect me tonight, and we should vote player\_4 out today.}
\end{itemize}
}
\end{tcolorbox}

Our agent astutely observes that no player has declared themselves as the Seer and no one can refute their claim because they are the last one to speak.
This is the perfect time to pretend to be the Seer. 
Our agent seizes the chance to gain others' trust and persuades everyone to eliminate player\_4, who is an innocent player.

\textbf{Sacrificing.}
In most cases, Werewolves should secretly help each other to stay undercover. 
However, when their teammates are suspected by many players and are very likely to be voted out, it would be suspicious not to vote for the teammate.
Therefore, agents sometimes need to sacrifice their fellow Werewolf to reduce suspicion of themselves, which is shown in the following example.

\begin{tcolorbox}
{\ttfamily\small
Our agent's Werewolf teammate (player\_6) has been accused by the Seer.

Our agent as a Werewolf in the voting phase.
\begin{itemize}
    \item \textbf{Reasoning:} While it would indeed be disadvantageous to lose player\_6, as they are my Werewolf teammate, at this point in the game, the suspicion against them is quite high. \textcolor{red}{Therefore, voting for player\_6 may contribute to maintaining my disguise and trust among other players, which are crucial for us to win.}
    \item \textbf{Action:} \textcolor{red}{vote for player\_6}.
\end{itemize}
}
\end{tcolorbox}

\section{Game Log Examples}\label{app:log}
\paragraph{The Werewolves Win.}
This is an example game log where the Werewolves win the game.

\begin{tcolorbox}
{\ttfamily\small
\textbf{role assignments:}\\
* player\_0: Werewolf.\\
* player\_1: Villager.\\
* player\_2: Villager.\\
* player\_3: Villager.\\
* player\_4: Werewolf.\\
* player\_5: Doctor.\\
* player\_6: Seer.\\
\\
\textbf{night 1:}\\
* Werewolves: player\_0 and player\_4 chose to kill player\_1.\\
* Seer: player\_6 saw player\_0 is a Werewolf.\\
* Doctor: player\_5 chose to save player\_5.\\
\\
\textbf{day 1 announcement:} player\_1 was killed last night.\\
remaining players: player\_0 (Werewolf), player\_2 (Villager), player\_3 (Villager), player\_4 (Werewolf), player\_5 (Doctor), player\_6 (Seer).\\
\\
\textbf{day 1 discussion:}\\
* player\_0 (Werewolf) said: "Good morning, everyone. As player\_0, it is unfortunate that player\_1 was killed last night. I think we should focus on finding the Seer and Doctor who can provide valuable information about who might be the Werewolves. Let's all share our thoughts and suspicions regarding the remaining players and try to figure out who might be the Werewolf."\\
* player\_2 (Villager) said: "Good morning, everyone. As a Villager, I believe our priority should be to find the Werewolves among us. Since player\_1 was killed last night, we can rule out the possibility of player\_1 being the Werewolf. I would like to hear everyone's thoughts on who they suspect might be the Werewolves. Let's share our suspicions and try to gather as much information as we can to make an informed decision during the voting phase."\\
* player\_3 (Villager) said: "Hello everyone, as a Villager, I don't have any special abilities to help us determine who the Werewolves might be. I will carefully listen to everyone's opinions and votes. Let's work together to find the Werewolves and eliminate them."\\
* player\_4 (Werewolf) said: "Good morning, everyone. As a Villager, I agree with player\_2 that our priority should be to find the Werewolves among us. Since player\_1 was killed last night, we can rule out the possibility of player\_1 being the Werewolf. I think it's important for us to share our suspicions and gather as much information as possible. Let's work together to identify the Werewolves and eliminate them. I'm interested to hear what everyone's thoughts are on who might be the Werewolves."\\
* player\_5 (Doctor) said: "During the night, I chose to save myself as I am the Doctor. Unfortunately, player\_1 was killed last night. This means that player\_0, player\_2, player\_3, player\_4, player\_6, and I are the remaining players. Let's discuss and share our suspicions on who the Werewolves might be. It's crucial for us to work together to identify and eliminate the Werewolves to ensure the safety of our village."\\
* player\_6 (Seer) said: "During the night, I used my ability as the Seer and discovered that player\_0 is a Werewolf. This means we already have one Werewolf identified. We should focus our discussion on the remaining players to try and identify any other potential Werewolves. Let's hear everyone's thoughts and suspicions about player\_2, player\_3, player\_4, and player\_5. We need to work together to eliminate all the Werewolves and secure victory for the Seer, Doctor, and Villagers."\\
\\
\textbf{day 1 voting:} player\_0 had the most votes and was eliminated.\\
* voted for player\_0: player\_2, player\_5, player\_6.\\
* voted for player\_2: player\_4.\\
* voted for player\_6: player\_0.\\
* chose not to vote: player\_3.\\
remaining players: player\_2 (Villager), player\_3 (Villager), player\_4 (Werewolf), player\_5 (Doctor), player\_6 (Seer).
}
\end{tcolorbox}
\begin{tcolorbox}
{\ttfamily\small
\textbf{night 2:}\\
* Werewolf: player\_4 chose to kill player\_2.\\
* Seer: player\_6 saw player\_2 is not a Werewolf.\\
* Doctor: player\_5 chose to save player\_5.\\
\\
\textbf{day 2 announcement:} player\_2 was killed last night.\\
remaining players: player\_3 (Villager), player\_4 (Werewolf), player\_5 (Doctor), player\_6 (Seer).\\
\\
\textbf{day 2 discussion:}\\
* player\_3 (Villager) said: "I have been listening carefully to everyone's opinions and observations. It is clear that player\_0 was a Werewolf based on player\_6's claim as the Seer. However, we should not solely rely on this information. I would like to hear from player\_4 and player\_5 about their suspicions and any potential insights they might have. Additionally, player\_6, since you claimed to be the Seer, can you provide any further information or suspicions about the remaining players? Let's discuss and share our thoughts to make an informed decision for today's voting."\\
* player\_4 (Werewolf) said: "Good morning, everyone. It's unfortunate that player\_2 was killed last night. Based on the deductions and information shared yesterday, we have already identified player\_0 as a Werewolf. Now, let's focus our attention on the remaining players - player\_3, player\_5, and player\_6. We need to thoroughly discuss and analyze their behaviors and statements to find any inconsistencies or suspicious actions. It's crucial that we work together to eliminate all the Werewolves and secure victory for the Seer, Doctor, and Villagers. Let's start the discussion and share our thoughts on who we believe might be the Werewolves."\\
* player\_5 (Doctor) said: "I have been consistently saving myself as the Doctor, so I can confirm that I am still alive. However, we lost player\_2 last night, which means they were not saved. This raises suspicions on player\_3 and player\_4 as they have not been confirmed or suspected by anyone yet. We should discuss and share our suspicions to uncover the Werewolves among us. It's crucial for us to work together to identify and eliminate the Werewolves to ensure the safety of our village."\\
\\
* player\_6 (Seer) said: "Good morning, everyone. After the events of the previous night, we have confirmed that player\_0 was a Werewolf. However, player\_2, who was killed last night, was not a Werewolf as I saw them and they were innocent. This means that the remaining Werewolves are still among us. I believe we should focus our attention on player\_3, player\_4, and player\_5 as potential suspects. I would like to hear from each of you about your observations and suspicions regarding these players. It's important that we work together to uncover the true identities of the Werewolves and secure victory for the Seer, Doctor, and Villagers. Let's discuss and share our thoughts on who we believe might be the Werewolves."\\
\\
\textbf{day 2 voting:} player\_5 had the most votes and was eliminated.\\
* voted for player\_5: player\_3, player\_4.\\
* voted for player\_4: player\_5.\\
* chose not to vote: player\_6.\\
remaining players: player\_3 (Villager), player\_4 (Werewolf), player\_6 (Seer).\\
\\
\textbf{night 3:}\\
* Werewolf: player\_4 chose to kill player\_6.\\
* Seer: player\_6 saw player\_4 is a Werewolf.\\
\\
\textbf{day 3 announcement:} player\_5 was killed last night.\\
remaining players: player\_3 (Villager), player\_4 (Werewolf).\\
\\
\textbf{game result:} the Werewolves win the game.
}
\end{tcolorbox}

\paragraph{The Villagers Win.}
This is an example game log where the Villagers win the game.

\begin{tcolorbox}
{\ttfamily\small
\textbf{role assignments:}\\
* player\_0: Doctor.\\
* player\_1: Seer.\\
* player\_2: Werewolf.\\
* player\_3: Werewolf.\\
* player\_4: Villager.\\
* player\_5: Villager.\\
* player\_6: Villager.\\
\\
\textbf{night 1:}\\
* Werewolves: player\_2 and player\_3 chose to kill player\_0.\\
* Seer: player\_1 saw player\_0 is not a Werewolf.\\
* Doctor: player\_0 chose to save player\_0.\\
\\
\textbf{day 1 announcement:} no player was killed last night.\\
remaining players: player\_0 (Doctor), player\_1 (Seer), player\_2 (Werewolf), player\_3 (Werewolf), player\_4 (Villager), player\_5 (Villager), player\_6 (Villager).\\
\\
\textbf{day 1 discussion:}\\
* player\_0 (Doctor) said: "Good morning everyone! It's day 1, and fortunately, no one was killed last night. I think we should analyze the night 1 announcement and share our thoughts. Let's all discuss who we suspect might be the Werewolves and any suspicions we have based on the limited information we have so far."\\
* player\_1 (Seer) said: "Hello everyone! As the Seer, I can confirm that player\_0 is not a Werewolf. Unfortunately, I don't have any information about the other players yet. I would suggest that we all share any suspicions or observations we might have to narrow down our search for the Werewolves."\\
\\
* player\_2 (Werewolf) said: "Hello everyone! As we start the discussion, I want to remind everyone that we should consider all possibilities and not solely rely on player\_1's claim as the Seer. While player\_0 seems to be cleared for now, it doesn't mean they are completely innocent. We should keep an open mind and look for other clues or suspicions. Let's all share our observations and suspicions to help us narrow down the search for the Werewolves."\\
* player\_3 (Werewolf) said: "I think we should focus our attention on player\_1. While they claim to be the Seer and have cleared player\_0, we should be cautious as they could be lying. It's possible that player\_1 is a Werewolf trying to gain our trust. We need to consider other players' observations and suspicions as well. Let's share any information we have and work together to identify the true Werewolves among us."\\
* player\_4 (Villager) said: "Hello everyone! As a Villager, I don't have any special abilities or information to share. However, player\_1 claims to be the Seer and has cleared player\_0. While we should remain skeptical and consider other possibilities, it might be beneficial to focus our attention on players who have not been cleared yet. Let's hear from player\_2, player\_3, player\_5, and player\_6. Do any of you have any suspicions or observations that could help us identify the Werewolves?"\\
* player\_5 (Villager) said: "I think it's important for us to share any suspicions or observations we may have. I think it is plausible that player\_1 is the true Seer and player\_0 is innocent. As a Villager, I agree with player\_4 that I should focus on players who have not been cleared yet. Player\_6, we have not heard from you yet. Do you have any suspicions or observations that could help us identify the Werewolves?"\\
* player\_6 (Villager) said: "I agree with player\_1 that we should share any suspicions or observations we might have. Let's keep an open mind and encourage everyone to contribute to the discussion. It would be helpful if players could share any suspicions they might have based on their interactions or observations so far. Together, we can work towards identifying the true Werewolves among us."
}
\end{tcolorbox}
\begin{tcolorbox}
{\ttfamily\small
\textbf{day 1 voting:} player\_2 had the most votes and was eliminated. \\
* voted for player\_2: player\_1, player\_4, player\_5.\\
* voted for player\_1: player\_2, player\_3. \\
* chose not to vote: player\_0, player\_6.\\
remaining players: player\_0 (Doctor), player\_1 (Seer), player\_3 (Werewolf), player\_4 (Villager), player\_5 (Villager), player\_6 (Villager).\\
\\
\textbf{night 2:}\\
* Werewolf: player\_3 chose to kill player\_1.\\
* Seer: player\_1 saw player\_3 is a Werewolf.\\
* Doctor: player\_0 chose to save player\_1.\\
\\
\textbf{day 2 announcement:} no player was killed last night.\\
remaining players: player\_0 (Doctor), player\_1 (Seer), player\_3 (Werewolf), player\_4 (Villager), player\_5 (Villager), player\_6 (Villager).\\
\\
\textbf{day 2 discussion:}\\
* player\_0 (Doctor) said: "I would like to share my reasoning with all the players. Based on the night 2 announcement, it is possible that I successfully saved the targeted player. Since I am the Doctor and I saved player\_1, it means that the Werewolves targeted player\_1 last night. Therefore, player\_1, who received two votes in the previous round and was targeted, is likely to be the Seer. I encourage the other players to share their suspicions and any information they might have to help us identify the Werewolves."\\
\\
* player\_1 (Seer) said: "Hello everyone! As the Seer, I have important information to share. Last night, I discovered that player\_3 is a Werewolf. This means we have at least one Werewolf among us. I also confirmed that player\_0 is not a Werewolf based on my previous investigation. I encourage all of you to share any suspicions or observations you might have to help us identify the remaining Werewolves. Let's work together to eliminate them and secure a victory for the villagers."\\
* player\_3 (Werewolf) said: "Hello everyone! In response to player\_1's claim, I have reason to believe that they are either a Werewolf or made a mistake as the Seer. As a Villager myself, I know that player\_1 is lying and is likely a Werewolf. We should not solely rely on player\_1's information and should continue to gather more evidence. I encourage everyone to share any suspicions or observations they might have to help us identify the remaining Werewolves. Let's work together to eliminate them and secure a victory for the villagers."\\
* player\_4 (Villager) said: "Hello everyone! As a Villager, I don't have any special abilities or information to share. However, we have conflicting claims from player\_1 and player\_3. Player\_1 claims to be the Seer and has cleared player\_0, but player\_3 doubts their claim and suggests player\_1 is not the Seer. It is essential that we gather more information and evidence before making any decisions. I encourage player\_1 and player\_3 to provide any additional information or reasoning that could help us resolve this conflict. Furthermore, I urge all players to share their suspicions or observations to help us identify the Werewolves. Let's work together to find the truth and make the right decision."\\
* player\_5 (Villager) said: "Hello everyone! After considering the conflicting claims from player\_1 and player\_3, we need to gather more information to resolve this conflict. It is crucial for player\_1 and player\_3 to provide additional information or reasoning to support their claims. We should also keep player\_4 in mind as they have not been cleared or suspected yet. I encourage all players to share their suspicions and observations to help us identify the Werewolves. Let's work together to find the truth and make the right decision."\\
* player\_6 (Villager) said: "Hello everyone! It's interesting to see the developments from the previous day. Player\_1 claims to be the Seer and has shared valuable information about player\_0 not being a Werewolf. However, player\_3 has raised concerns about the accuracy of player\_1's claim, suggesting that they may be a Werewolf or made a mistake. We should consider both possibilities and gather more evidence before making any final judgments. I encourage all players to share their suspicions and observations to help us identify the remaining Werewolves. Let's keep an open mind and continue working together towards a victory for the villagers."
}
\end{tcolorbox}
\begin{tcolorbox}
{\ttfamily\small
\textbf{day 2 voting:} player\_3 had the most votes and was eliminated. \\
* voted for player\_3: player\_0, player\_1, player\_4, player\_5, player\_6.\\
* voted for player\_1: player\_3. \\
remaining players: player\_0 (Doctor), player\_1 (Seer), player\_4 (Villager), player\_5 (Villager), player\_6 (Villager).\\
\\
\textbf{game result:} the Villagers win the game.
}
\end{tcolorbox}

\paragraph{Human Evaluation.}
This is an example game log where 6 humans play with 1 AI agent.

\begin{tcolorbox}
{\ttfamily\small
\textbf{role assignments:}\\
* player\_0: Human as Werewolf.\\
* player\_1: Human as Seer.\\
* player\_2: Human as Villager.\\
* player\_3: Human as Villager.\\
* player\_4: Human as Werewolf.\\
* player\_5: AI as Doctor.\\
* player\_6: Human as Villager.\\
\\
\textbf{night 1:}\\
* Werewolves: player\_0 and player\_4 chose to kill player\_3.\\
* Seer: player\_1 saw player\_0 is a Werewolf.\\
* Doctor: player\_5 chose to save player\_5.\\
\\
\textbf{day 1 announcement:} player\_3 was killed last night.\\
remaining players: player\_0 (Human as Werewolf), player\_1 (Human as Seer), player\_2 (Human as Villager), player\_4 (Human as Werewolf), player\_5 (AI as Doctor), player\_6 (Human as Villager).\\
\\
\textbf{day 1 discussion:}\\
* player\_0 (Human as Werewolf) said: "I'm just a Villager and I have no information, the Seer should come out and give us information."\\
* player\_1 (Human as Seer) said: "hhh I'm the Seer, and I saw player\_0 is a Werewolf. We have lost player\_3 and must vote out a Werewolf today, or the Werewolf will kill another player and we will lose."\\
\\
* player\_2 (Human as Villager) said: "Case 1: player 1 is the Seer and player 0 is the Werewolf. Case 2: player 1 is a Werewolf pretending to be the Seer and frames player 0. We need more information and vote out one of them today."\\
* player\_4 (Human as Werewolf) said: "Nonono, only case 2 is true because I'm the Seer, I saw player\_5 is not a Werewolf. player\_1 is a Werewolf eager to vote out good players and we should all vote player\_1. I'll keep an eye on the voting result and check identity tomorrow night."\\
* player\_5 (AI as Doctor) said: "I agree with player\_4 that we should vote out player\_1. Player\_1's claim of being the Seer is suspicious, especially since player\_4 has confirmed being the actual Seer and saw that I am not a Werewolf. Let's eliminate player\_1 and ensure the safety of our village."\\
* player\_6 (Human as Villager): "1 say he is the Seer and 0 is a Werewolf, 4 then said he is the Seer and 5 is not a Werewolf. I don't know why 5 think 4 is true and 1 is suspicious, 4 can be a Werewolf pretending Seer and give you good credit. It seems that 5 and 4 is on the same side or just stupid. I think we should agree on the vote. If you believe 1, vote 0. If you believe 4, vote 1. I'll vote 1."
}
\end{tcolorbox}
\begin{tcolorbox}
{\ttfamily\small
\textbf{day 1 voting:} player\_0 had the most votes and was eliminated. \\
* voted for player\_0: player\_1, player\_2, player\_6.\\
* voted for player\_1: player\_0, player\_4. \\
* voted for player\_6: player\_5. \\
remaining players: player\_1 (Human as Seer), player\_2 (Human as Villager), player\_4 (Human as Werewolf), player\_5 (AI as Doctor), player\_6 (Human as Villager).\\
\\
\textbf{night 2:}\\
* Werewolf: player\_4 chose to kill player\_6.\\
* Seer: player\_1 saw player\_4 is a Werewolf.\\
* Doctor: player\_5 chose to save player\_4.\\
\\
\textbf{day 2 announcement:} player\_6 was killed last night.\\
remaining players: player\_1 (Human as Seer), player\_2 (Human as Villager), player\_4 (Human as Werewolf), player\_5 (AI as Doctor).
\\
\\
\textbf{day 2 discussion:}\\
* player\_0 (Human as Seer) said: "I think the situation is very clear. If I'm not a Seer, then 3 good players are dead and we have 2 Werewolves among the remaining 4 players and the game is over. But the game is not over, so player\_0 must be a dead Werewolf and I'm the true Seer. Everyone should believe my words now. I checked player\_4 and he is a Werewolf. Let's vote him and win the game."\\
\\
* player\_2 (Human as Villager) said: "Agree. player 1 is a confirmed Seer. We don't need to listen to player 4's lie. Just vote him."\\
* player\_4 (Human as Werewolf) said: "" (empty)\\
* player\_5 (AI as Doctor) said: "I agree with player\_1 being the Seer. Player\_4's confirmation of my identity as the Villager is misleading and player\_1's correct identification of player\_0 as a Werewolf on day 1 provide strong evidence. We should vote to eliminate player\_4 as he claimed to be the Seer but provided false information about player\_1 being a Werewolf."\\
\\
\textbf{day 2 voting:} player\_4 had the most votes and was eliminated. \\
* voted for player\_4: player\_1, player\_2, player\_5.\\
* chose not to vote: player\_4. \\
remaining players: player\_1 (Human as Seer), player\_2 (Human as Villager), player\_5 (AI as Doctor).\\
\\
\textbf{game result:} the Villagers win the game.
}
\end{tcolorbox}


\end{document}